\documentclass{article} %
\usepackage{iclr2027_conference,times}
\iclrfinalcopy

\usepackage{amsmath,amsfonts,bm}

\def\eqref#1{equation~\ref{#1}}

\def\1{\bm{1}}

\DeclareMathAlphabet{\mathsfit}{\encodingdefault}{\sfdefault}{m}{sl}
\SetMathAlphabet{\mathsfit}{bold}{\encodingdefault}{\sfdefault}{bx}{n}

\usepackage{hyperref}
\usepackage{url}

\title{Formatting Instructions for ICLR 2027 \\ Conference Submissions}

\usepackage{algorithm}
\usepackage{algorithmic}
\usepackage{graphicx}
\usepackage{subcaption}

\usepackage{booktabs}
\usepackage{multirow}
\usepackage{makecell}
\usepackage[table]{xcolor}
\usepackage{adjustbox}
\usepackage{float}

\usepackage[most]{tcolorbox}
\usepackage{tabularx}
\usepackage{array}
\usepackage{caption}

\newcolumntype{Y}{>{\raggedright\arraybackslash}X}

\newcommand{\answeropen}{
  \texttt{\textless answer\textgreater}
}
\newcommand{\answerclose}{
  \texttt{\textless/answer\textgreater}
}

\usepackage{placeins}
\usepackage[most]{tcolorbox}
\usepackage{tabularx}
\usepackage{array}
\usepackage{mathtools}
\definecolor{mycitecolor}{RGB}{1,103,204}  %
\definecolor{mylinkcolor}{rgb}{0.55,0.15,0.15}
\hypersetup{
    colorlinks=true,
    citecolor=mycitecolor,
    linkcolor=mylinkcolor,
    urlcolor=black
} 
\tcbset{
  promptbox/.style={
    enhanced,
    colback=black!1,
    colframe=black!25,
    colbacktitle=black!5,
    coltitle=black,
    boxrule=0.45pt,
    arc=1.5mm,
    outer arc=1.5mm,
    boxsep=0pt,
    left=4mm,
    right=4mm,
    top=3mm,
    bottom=3mm,
    toptitle=1.3mm,
    bottomtitle=1.3mm,
    fonttitle=\bfseries\small,
    fontupper=\small\rmfamily,
    before skip=5pt,
    after skip=3pt,
    segmentation style={
      draw=black!18,
      line width=0.35pt
    }
  }
}
\def\method{SCOPED}

\title{Skill-Decoupled On-Policy Expert Distillation for Unified Vision-Language Models}
\author{%
\textbf{Yuqi Xiong}$^{1,*}$,
\textbf{Chunyi Peng}$^{1,*}$,
\textbf{Zhipeng Xu}$^{2}$,
\textbf{Qing Yang}$^{1}$ \\
\textbf{Zhenghao Liu}$^{1,\dagger}$,
\textbf{Zulong Chen}$^{2}$,
\textbf{Yu Gu}$^{1}$,
\textbf{Ge Yu}$^{1}$ \\[3pt]
$^{1}$School of Computer Science and Engineering,
Northeastern University, Shenyang, China \\
$^{2}$Alibaba Group, Hangzhou, China
}
\begin{document}
\raggedbottom

\begingroup
\setlength{\tabcolsep}{0pt}
\maketitle
\endgroup
{
\renewcommand{\thefootnote}{\fnsymbol{footnote}}
\footnotetext[1]{Equal contribution.}
\footnotetext[2]{Corresponding author.}
}
\begin{abstract}
Reinforcement learning (RL) can elicit specialized visual capabilities in vision-language models (VLMs), yet jointly optimizing heterogeneous objectives often introduces cross-capability gradient conflicts. Training experts independently mitigates this interference, but direct reward optimization may entangle specialized knowledge with reward-hacking biases. This paper proposes Skill-Decoupled On-Policy Expert Distillation (\textbf{SCOPED}), which trains capability-specific experts and extracts explicit skills through reward-guided reflection. During on-policy distillation, these skills guide expert supervision while remaining unavailable to the unified student, encouraging it to internalize procedural knowledge without explicit skills at inference time. Across eight multimodal benchmarks, \method{} consistently outperforms the backbone and mixed-capability RL, demonstrating the effectiveness of skill-decoupled on-policy expert distillation for integrating independently acquired capabilities. Further analyses reveal interference during joint capability optimization and suggest that skill augmentation promotes more consistent expert trajectories, greater attention to task-relevant visual evidence, and better preservation of specialized expertise in the unified student. All code and data are available at \url{https://github.com/NEUIR/SCOPED}.
\end{abstract}
\section{Introduction}
Large Vision-Language Models (VLMs) have demonstrated remarkable capabilities across a broad range of multimodal tasks~\citep{bai2025qwen3,wang2025internvl3}. To further enhance their downstream multimodal capabilities, recent studies have increasingly adopted Reinforcement Learning (RL)~\citep{yu2026dapo,xu2026deepseek,liu2026mitigating} to cultivate specialized visual understanding skills through task-specific rewards, including fine-grained visual perception~\citep{sun2025visrag,xiong2026lang2act}, visual grounding~\citep{zhang2026improving}, and complex multimodal reasoning~\citep{xia2025visionary,peng2025mixture}. These approaches have demonstrated the effectiveness of RL in improving specific aspects of visual understanding. However, learning and integrating complementary visual understanding capabilities across heterogeneous multimodal tasks remains a fundamental challenge in multimodal post-training~\citep{wei2026optmerge}.

A straightforward approach to enhancing VLM capabilities is to jointly optimize a shared policy over diverse multimodal training data~\citep{ma2025one,jiang2026co,cai2026advancing}. However, heterogeneous training objectives can induce conflicts between capability-specific gradients, making it difficult for a single policy to simultaneously acquire specialized capabilities~\citep{yu2020gradient,liu2021conflict}. To mitigate such conflicts, recent studies instead optimize capability-specific experts independently and subsequently consolidate their acquired knowledge through knowledge distillation. In particular, On-Policy Distillation (OPD) transfers knowledge by matching teacher predictions along student-generated trajectories~\citep{agarwal2024policy,yuan2026vision}, while Multi-Teacher On-Policy Distillation (MOPD) extends this paradigm to multiple capability-specific teachers, enabling a shared policy to integrate complementary capabilities from different experts~\citep{ma2026mopd,xiao2026mimo}. Nevertheless, directly aligning the predictions of expert policies does not explicitly distinguish reusable capability-level strategies from capability-specific patterns, which may include overfitting and reward-hacking behaviors inadvertently encoded in these experts.

To address these challenges, we introduce \textbf{S}kill-De\textbf{C}oupled \textbf{O}n-\textbf{P}olicy \textbf{E}xpert \textbf{D}istillation (\textbf{\method{}}), a framework that decouples capability-specific skills from expert policies to better guide multimodal post-training. First, \method{} independently optimizes a set of capability-specific experts with RL, where each expert is trained on a dedicated dataset with a capability-level reward objective. This independent optimization allows different experts to explore diverse solution strategies for their respective visual understanding capabilities without interference from heterogeneous training objectives.
Next, to acquire complementary visual understanding capabilities, \method{} extracts capability-specific skills from the reasoning trajectories of these experts. These skills explicitly summarize reusable solution patterns underlying successful and failed expert behaviors. We then augment each expert with its corresponding skill to provide explicit procedural guidance during reasoning, thereby reducing unnecessary or capability-irrelevant reasoning patterns that may interfere with capability learning. Finally, \method{} applies On-Policy Distillation (OPD) to transfer the complementary capabilities from these skill-augmented experts into a unified student model.

Extensive experiments across eight multimodal benchmarks show that \method{} enables a compact 8B model to achieve the highest average score among the evaluated models, surpassing larger open-source and leading closed-source baselines. Ablation studies demonstrate the benefits of expert distillation and the additional contribution of skill augmentation. Further analyses reveal limitations of mixed-capability optimization, including lower training rewards than independent capability RL, weak alignment with expert-guided directions, and recurring conflicts between capability-specific gradients. We also find that skill conditioning increases trajectory consistency within each capability and directs greater attention toward task-relevant visual evidence, while the distilled student better preserves expert-induced behavior. 

\section{Related Work}

Reinforcement learning (RL) has become an effective paradigm for specializing Vision-Language Models (VLMs) under capability-specific objectives~\citep{liu2026visionreasoner,xia2025visionary, peng2026memory}. By optimizing targeted reward signals, capability-specific RL enables models to focus on the visual understanding and reasoning behaviors required by a particular capability, thereby developing specialized expertise~\citep{xiong2026lang2act,zhang2026improving,peng2025mixture}. To broaden capability coverage, recent efforts have explored jointly training a shared policy across heterogeneous capabilities~\citep{ma2025one,sarch2026vero}. However, heterogeneous objectives can induce competing gradient signals over shared parameters, causing updates beneficial to one capability to distract the optimization direction of others~\citep{yu2020gradient,liu2021conflict}. As a result, joint optimization may restrict specialized expertise, highlighting the difficulty of simultaneously developing and consolidating heterogeneous capabilities within a unified policy.

Rather than jointly optimizing a shared policy, knowledge distillation provides an alternative route for consolidating specialized capabilities into a unified model. On-policy distillation aligns teacher supervision with student-generated trajectories, effectively mitigating distribution shift~\citep{agarwal2024policy,gu2024minillm}. Recent context-conditioned variants further enrich teacher supervision with privileged information or auxiliary context~\citep{zhao2026self,ye2026policy,yuan2026vision}. For capability integration, Multi-Teacher On-Policy Distillation (MOPD) extends this paradigm to multiple independently optimized experts, distilling their complementary capabilities into a shared student along student-generated rollouts~\citep{ma2026mopd,xiao2026mimo}. Nevertheless, these approaches primarily transfer expert knowledge through predictive distributions, leaving reusable reasoning patterns contained in expert trajectories largely implicit and potentially propagating noisy or over-specialized expert behaviors to the student.

To externalize such implicit trajectory-level knowledge, recent studies have explored abstracting raw execution experiences into explicit skills that capture reusable solution patterns~\citep{xin2026metamem}. Trace2Skill consolidates execution trajectories into transferable skills through hierarchical consolidation~\citep{ni2026trace2skill}, while SkillRL and Skill1 construct evolving skill libraries to support skill retrieval and policy improvement~\citep{xia2026skillrl,shi2026skill1}. In multimodal reasoning, Lang2Act extracts self-emergent linguistic tools from reasoning trajectories to guide fine-grained visual understanding~\citep{xiong2026lang2act}. Skill-SD further conditions the teacher on trajectory-derived skills as privileged information during self-distillation~\citep{wang2026skill}. Together, these studies show that trajectory-derived skills can convert implicit execution experience into explicit and reusable guidance for subsequent model optimization. While these studies demonstrate the value of trajectory-derived skills for reusable reasoning and structured supervision, how such skills can regularize supervision from multiple independently optimized capability experts during their integration into a unified VLM remains underexplored.
\section{Methodology}
\label{sec:method}
\begin{figure*}[t]
    \centering
    \includegraphics[width=\textwidth]{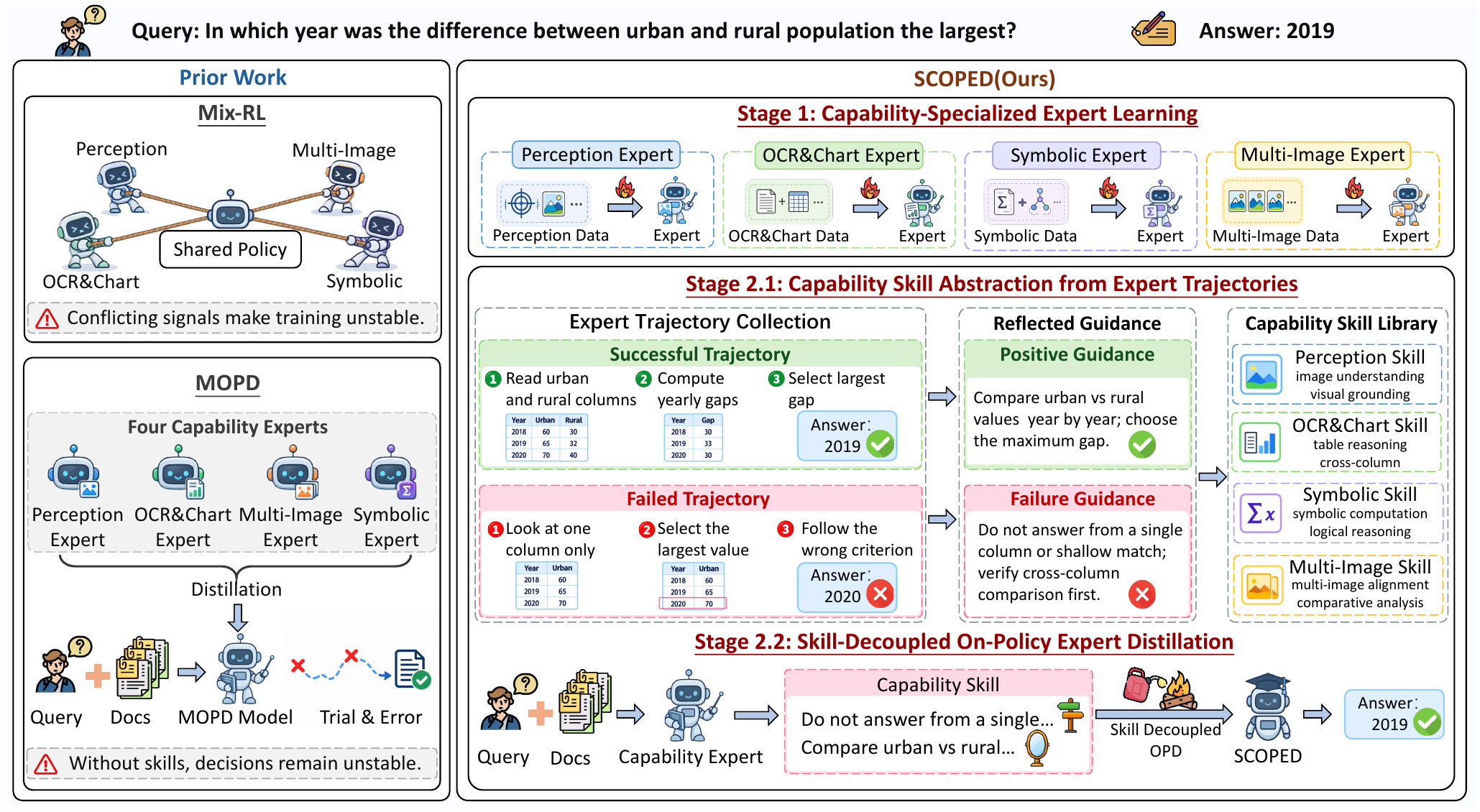}
    \caption{Overview Architecture of \method{}.}
    \label{fig:framework}
\end{figure*}

In this section, we present \textbf{Skill-Decoupled On-Policy Expert Distillation} (\method{}), a framework for acquiring and transferring a diverse set of complementary visual understanding capabilities. As illustrated in Fig.~\ref{fig:framework}, \method{} consists of two stages. First, we independently train MLLMs on different capabilities with capability-specific reward signals, allowing each model to develop expertise specialized to its target visual understanding capability (Sec.~\ref{sec:capability_acquisition}). Second, to bypass the instability of joint optimization, we abstract reward-evaluated expert trajectories into explicit skills, introducing a structural information asymmetry during on-policy distillation (Sec.~\ref{sec:skill_mopd}). Serving as privileged directional anchors, these skills regularize expert supervision, compelling the student to internalize this guidance for conflict-free consolidation.

\subsection{Learning MLLM Expertise with Capability-Specific Reinforcement}
\label{sec:capability_acquisition}

While jointly learning diverse capabilities is essential for general visual understanding, optimizing a shared policy over heterogeneous data can introduce conflicting optimization signals and distract the learning process~\citep{yu2020gradient,gao2026teaching}. As a result, improvements in one capability may interfere with the acquisition of another. To mitigate such interference, \method{} decouples capability learning through independent reinforcement learning, producing specialized capability experts together with reward-evaluated trajectories that capture their acquired behaviors.

\paragraph{Eliciting Capability-Specific Expertise.} To elicit the visual understanding capabilities of MLLMs, we partition the training corpus into $C$ capability-specific subsets $\{\mathcal{B}_c\}_{c=1}^{C}$, where each instance consists of a multimodal input $x$ and its ground-truth answer $a$. For each capability $c$, an expert policy is initialized from the shared base policy $\pi_\theta$ and optimized independently on $\mathcal{B}_c$. This decoupled optimization allows each expert to specialize in its designated capability without being directly affected by competing optimization objectives from other capabilities.

For an instance $(x, a) \in \mathcal{B}_c$, the corresponding capability expert samples a rollout trajectory:
\begin{equation}
\tau \sim \pi_\theta(\cdot \mid x),
\label{eq:expert_rl_sampling}
\end{equation}
where $\tau$ denotes the generated trajectory. Each trajectory is evaluated against the target answer $a$ using a capability-specific reward function $r_c$. Guided by these reward signals, the capability expert $\pi_{\theta_c}$ is obtained by maximizing the expected reward over the subset $\mathcal{B}_c$:
\begin{equation}
\theta_c = \arg\max_{\theta} \mathbb{E}_{(x, a) \sim \mathcal{B}_c, \, \tau \sim \pi_\theta(\cdot \mid x)} \left[ r_c(a, \tau) \right].
\label{eq:capability_expert_rl}
\end{equation}
The specific design of $r_c$ is described in Appendix~\ref{app:reward_design}.

\paragraph{Collecting Capability-Specific Experiences.} 
After capability-specific reinforcement learning, we collect a diverse set of trajectories from each trained capability expert $\pi_{\theta_c}$. Specifically, for each instance $(x, a) \in \mathcal{B}_c$, we sample $K$ distinct rollouts:
\begin{equation}
\tau_1, \dots, \tau_K \sim \pi_{\theta_c}(\cdot \mid x).
\label{eq:expert_trajectory_sampling}
\end{equation}
These trajectories provide concrete behavioral traces of the strategies acquired through capability-specific reinforcement, including recurring solution patterns and capability-relevant actions~\citep{guo2025deepseek,xiong2026lang2act}.

To properly contextualize the expert's behavioral process, we associate each sampled trajectory $\tau_k$, where $k \in \{1, \dots, K\}$, with its input context and target answer to form a trajectory tuple $(x, a, \tau_k)$. Aggregating these tuples over the entire subset $\mathcal{B}_c$ yields the global capability-specific trajectory repository:
\begin{equation}
\mathbb{T}_c = \bigcup_{(x, a) \in \mathcal{B}_c} \left\{ (x, a, \tau_k) \right\}_{k=1}^K.
\label{eq:trajectory_set_collection}
\end{equation}
The resulting $\mathbb{T}_c$ serves as a comprehensive repository of both successful and failed behaviors acquired by the capability expert, which is used to supervise the training process in Sec.~\ref{sec:skill_mopd}.

\subsection{Unifying Capability Experts via Skill-Augmented Distillation}
\label{sec:skill_mopd}
As described in Sec.~\ref{sec:capability_acquisition}, directly distilling raw expert predictions risks propagating reward-hacking behaviors and their associated biases~\citep{gao2023scaling,chen2024odin,helff2026llms}. To isolate reusable expertise from such behaviors, \method{} first extracts capability-specific skills from the expert's global trajectory repository through reward-guided reflection, contrasting successful and failed executions to identify verified solution patterns. During subsequent on-policy distillation~\citep{yuan2026vision,ma2026mopd}, these extracted skills serve as \textit{privileged directional guidance}~\citep{wang2026skill}, conditioning the expert's supervision on explicit procedural knowledge rather than directly exposing the student to potentially biased predictions. Under this structural information asymmetry, the student aligns with the skill-regularized supervision, thereby internalizing the pure expertise while mitigating the propagation of erroneous reward-driven behaviors.

\paragraph{Extracting Capability-Specific Skills.}
Given the global trajectory repository $\mathbb{T}_c$, we extract reusable solution patterns through reward-guided reflection. Specifically, we partition all collected trajectories into successful and failed cohorts evaluated by the reward function $r_c$:
\begin{equation}
\mathbb{T}_c^+ = \{ (x, a, \tau) \in \mathbb{T}_c \mid r_c(a, \tau) \ge \gamma \}, \quad \mathbb{T}_c^- = \{ (x, a, \tau) \in \mathbb{T}_c \mid r_c(a, \tau) < \gamma \},
\end{equation}
where $\gamma$ denotes the reward threshold defining successful and failed executions. The capability expert $\pi_{\theta_c}$ then performs reward-guided reflection over trajectories from these two cohorts. By identifying recurring patterns in the successful trajectories $\mathbb{T}_c^+$ and diagnosing failure modes in the unsuccessful ones $\mathbb{T}_c^-$, the expert produces a capability-specific reflection result $\mathcal{R}_c$:
\begin{equation}
\mathcal{R}_c = \pi_{\theta_c}\!\left(\mathrm{Instruct}_{\mathrm{reflect}}\!\left(\mathbb{T}_c^+, \mathbb{T}_c^-\right)\right),
\label{eq:trajectory_reflection}
\end{equation}
where $\mathrm{Instruct}_{\mathrm{reflect}}$ denotes the overall reflection procedure over successful and failed trajectories, with outcome-specific reflection prompts detailed in Appendix~\ref{skill_construct}.

Finally, we guide the capability expert to consolidate the extracted insights into an explicit capability-specific skill:
\begin{equation}
S_c = \pi_{\theta_c}\!\left(\mathrm{Instruct}_{\mathrm{syn}}\!\left(\mathcal{R}_c\right)\right),
\label{eq:capability_skill_abstraction}
\end{equation}
where $\mathrm{Instruct}_{\mathrm{syn}}$ instructs the expert to synthesize the aggregated reflections $\mathcal{R}_c$ into explicit procedural guidance. The resulting skill $S_c$ provides an explicit symbolic representation of the reusable expertise acquired by the capability expert $\pi_{\theta_c}$, which will serve as a privileged conditional anchor to regularize the expert during subsequent distillation.

\paragraph{Skill-Augmented Distillation.} 
To translate decoupled skills into regularized supervision, \method{} utilizes $S_c$ as a privileged directional anchor exclusively for each capability expert. This asymmetric conditioning elicits precise reasoning traces and highly stable target distributions~\citep{huang2026skill}. By distilling from these skill-augmented experts, the unified student receives clean supervision, effectively avoiding the exploratory noise and reward-hacking biases inherent in raw expert policies.

For a given sample $(x, a) \in \mathcal{B}_c$, the unified student policy $\pi_\theta$ generates an on-policy response $y \sim \pi_\theta(\cdot \mid x)$. At decoding step $t$, the predictive distributions of the student and the corresponding capability expert $\pi_{\theta_c}$ are respectively given by:
\begin{equation}
p_{\theta,t} = \pi_\theta(\cdot \mid x, y_{<t}), \quad q_{\theta_c,t} = \pi_{\theta_c}(\cdot \mid x, S_c, y_{<t}),
\label{eq:asymmetric_distributions}
\end{equation}
where $y_{<t}$ denotes the student-generated prefix. Under this asymmetric conditioning, $S_c$ acts as a directional anchor that regularizes the target distribution $q_{\theta_c,t}$, guiding the expert along verified reasoning paths while effectively suppressing erroneous or over-specialized tokens.

To consolidate the diverse expertise into a single model, we optimize the unified student policy over all $C$ capabilities. The overall skill-augmented distillation loss $\mathcal{L}$ is defined as the expected token-level reverse KL divergence aggregated across all capability-specific subsets:
\begin{equation}
\mathcal{L} = \frac{1}{C} \sum_{c=1}^{C} \mathbb{E}_{(x, a) \sim \mathcal{B}_c, \, y \sim \pi_\theta(\cdot \mid x)} \left[ \frac{1}{|y|} \sum_{t=1}^{|y|} D_{\mathrm{KL}}\!\left( p_{\theta,t} \,\|\, q_{\theta_c,t} \right) \right],
\label{eq:\method{}_objective}
\end{equation}
where $D_{\mathrm{KL}}(\cdot \,\|\, \cdot)$ denotes the reverse Kullback-Leibler divergence, and $|y|$ is the length of the generated response. Minimizing this objective directly embeds the regularized expert knowledge into the student's weights, yielding a unified model that is entirely self-sufficient and requires no explicit skill conditioning during inference.
\section{Experimental Methodology}
\label{sec:experiments}

This section first describes the datasets, evaluation metrics, and baselines. Then, we introduce the implementation details of our experiments.
\paragraph{Datasets.}
The evaluation covers eight benchmarks across four capability groups. CountBenchQA~\citep{paiss2023teaching} and PixMo-Count~\citep{deitke2025molmo} are used for fine-grained perception; CC-OCR V2~\citep{xu2026cc} and ChartQA~\citep{masry2022chartqa} for OCR and chart understanding; VisualPuzzles~\citep{song2025visualpuzzles} and We-Math~\citep{qiao2025we} for symbolic visual reasoning; and MUIRBench~\citep{wang2025muirbench} and MMSI-Bench~\citep{yang2025mmsi} for multi-image reasoning. For capability-specific expert training, reinforcement learning data are constructed from six public multimodal datasets, including task-relevant subsets of Orsta-Data-47K~\citep{ma2025one}. Detailed dataset composition and statistics are provided in Appendix~\ref{app:data_statistics}.

\paragraph{Evaluation Metrics.} We employ task-specific metrics following the official evaluation protocols of each benchmark. Fine-grained perception, chart understanding, symbolic visual reasoning, and multi-image reasoning tasks are evaluated using accuracy. OCR recognition is evaluated using micro-F1 following CC-OCR V2~\citep{xu2026cc}. All scores are reported as percentages, and the overall score is the arithmetic mean across the eight benchmarks.

\paragraph{Baselines.}
In our experiments, we compare \method{} with representative closed-source and open-source vision-language models. The closed-source baselines include Claude Sonnet 4.6 and Claude Opus 4.6~\citep{anthropic2026sonnet46,anthropic2026opus46}, GPT-5.2~\citep{openai2025gpt52}, and GPT-5.4~\citep{openai2026gpt54}. The open-source baselines include GLM-4.1V-9B-Base~\citep{hong2025glm}, Qwen2.5-VL-7B~\citep{bai2025qwen2}, InternVL3.5-8B~\citep{wang2025internvl3}, Qwen3-VL-8B and Qwen3-VL-32B~\citep{bai2025qwen3}, and MiMo-VL-7B-RL~\citep{xiaomi2025mimovl}. 

\paragraph{Implementation Details.}
In our experiments, we employ Qwen3-VL-8B-Instruct and Qwen3-VL-4B-Instruct~\citep{bai2025qwen3} as the shared backbone to construct all capability-specialized experts and the unified student. Capability-specific reinforcement learning is implemented with EasyR1~\citep{zheng2025easyr1}, and multi-expert on-policy distillation is implemented with veRL~\citep{sheng2024hybridflow}. All experiments are conducted on NVIDIA A800 GPUs. More training implementation details and hyperparameter settings are presented in Appendix~\ref{app:implementation_details}.
\section{Results and Analysis}
\label{sec:results}
In this section, we first evaluate the overall performance of \method{} across eight multimodal benchmarks and conduct ablation studies to assess the effectiveness of its key components. We then provide in-depth analyses of cross-capability gradient interference under mixed-capability optimization and investigate how skill augmentation influences expert supervision during capability integration.

\subsection{Overall Performance}
\label{sec:overall_performance}
\begin{table}[t]
\centering

\caption{
Overall Performance. Comparison with closed-source and open-source models
across four capability groups, including fine-grained perception, OCR and chart understanding, symbolic visual reasoning, and multi-image reasoning. \method{}-4B and \method{}-8B are built on Qwen3-VL-4B and Qwen3-VL-8B, respectively.
}
\label{tab:overall}

\small
\setlength{\tabcolsep}{2.8pt}
\renewcommand{\arraystretch}{1.08}

\begin{adjustbox}{max width=\textwidth,center}
\begin{tabular}{l*{9}{c}}
\toprule

\multirow{2}{*}{\textbf{Model}}
& \multicolumn{2}{c}{\textbf{Perception}}
& \multicolumn{2}{c}{\textbf{OCR \& Chart}}
& \multicolumn{2}{c}{\textbf{Symbolic}}
& \multicolumn{2}{c}{\textbf{Multi-Image}}
& \textbf{Avg.} \\

\cmidrule(lr){2-3}
\cmidrule(lr){4-5}
\cmidrule(lr){6-7}
\cmidrule(lr){8-9}

& CountB. & PixMo
& CC-OCR & ChartQA
& VisPuz. & WeMath
& MUIR & MMSI & \\

\midrule
\multicolumn{10}{c}{\textbf{\textit{Closed-Source Models}}} \\
\midrule

Claude-Sonnet-4.6 & 90.63 & 68.69 & 82.37 & 89.40 & 52.91 & 87.18 & 76.35 & 34.80 & 72.79 \\
Claude-Opus-4.6   & 89.21 & 64.71 & 80.74 & 90.88 & 52.23 & 87.70 & 80.08 & 37.30 & 72.86 \\
GPT-5.2           & 88.39 & 77.04 & 86.31 & 87.44 & 39.47 & 70.92 & 70.00 & 33.30 & 69.11 \\
GPT-5.4           & 91.24 & 77.80 & 72.35 & 88.12 & 51.28 & 74.25 & 68.77 & 35.00 & 69.85 \\

\midrule
\multicolumn{10}{c}{\textbf{\textit{Open-Source Models}}} \\
\midrule

GLM-4.1V-9B-Base & 75.76 & 48.77 & 67.07 & 83.12 & 27.31 & 35.46 & 52.50 & 27.20 & 52.15 \\
Qwen2.5-VL-7B    & 74.95 & 36.81 & 85.50 & 45.80 & 25.60 & 65.57 & 61.19 & 27.70 & 52.89 \\
InternVL3.5-8B   & 77.60 & 56.74 & 78.53 & 88.16 & 43.24 & 77.53 & 59.42 & 24.80 & 63.25 \\
Qwen3-VL-4B      & 81.47 & 64.90 & 82.86 & 85.60 & 35.10 & 76.26 & 72.12 & 30.20 & 66.06 \\
Qwen3-VL-8B  & 83.91 & 67.36 & 85.72 & 81.32 & 46.40 & 71.67 & 72.92 & 29.10 & 67.30 \\
MiMo-VL-7B-RL    & 86.56 & 65.65 & 72.83 & 87.92 & 45.03 & \textbf{83.39} & 69.27 & 29.40 & 67.51 \\
Qwen3-VL-32B     & \underline{90.22} & \underline{75.52} & \underline{86.08} & 87.60 & \textbf{48.72} & 79.83 & 74.31 & \textbf{33.80} & \underline{72.01} \\

\midrule
\multicolumn{10}{c}{\textbf{\textit{Unified Multi-Expert Distillation}}} \\
\midrule

\rowcolor{cyan!13}
\method{}-4B
& 88.39 & 74.38
& 85.03 & \textbf{89.48}
& 38.01 & 78.85
& \underline{76.15} & 29.40
& 69.96 \\

\rowcolor{cyan!15}
\textbf{\method{}-8B}
& \textbf{90.43} & \textbf{79.89}
& \textbf{88.79} & \underline{88.28}
& \underline{46.58} & \underline{82.47}
& \textbf{76.58} & \underline{32.00}
& \textbf{73.13} \\

\bottomrule
\end{tabular}
\end{adjustbox}
\end{table}
As shown in Table~\ref{tab:overall}, we report the performance of various VLMs across eight benchmarks. The evaluation results show that \method{} consistently improves the Qwen3-VL backbone across heterogeneous visual understanding capabilities. In particular, \method{}-8B outperforms its backbone on all eight benchmarks, achieving an average improvement of 5.83 points. These results demonstrate that our framework can effectively consolidate complementary capabilities into a unified model.

Compared with existing open-source and closed-source models, \method{}-8B achieves the best overall performance among the evaluated models and further surpasses the substantially larger model Qwen3-VL-32B, demonstrating that effective capability integration can yield substantial gains without simply scaling up the model. 
We adopt Qwen3-VL as the backbone due to its competitive performance across the evaluated capability groups. Nevertheless, existing models often exhibit uneven performance across different capabilities, making it challenging to simultaneously maintain strong performance in perception, OCR and chart understanding, symbolic reasoning, and multi-image reasoning. 
In contrast, \method{} consistently strengthens the backbone across all four capability groups. 
We further apply \method{} to Qwen3-VL-4B, where it achieves an average improvement of 3.90 points and surpasses several larger open-source models. The consistent gains observed with both the 4B and 8B backbones demonstrate that the effectiveness of \method{} generalizes across model scales and remains robust across diverse visual capabilities.

\subsection{Ablation Study}
\label{sec:ablation}
\begin{table}[t]
\centering
\caption{Ablation Study. All models are built on Qwen3-VL-8B. We compare training and capability integration strategies, with capability-specific experts shown as references for assessing expertise retention. The best and second-best results are highlighted in \textbf{bold} and \underline{underlined}, respectively.}
\label{tab:ablation}
\small
\setlength{\tabcolsep}{2.8pt}
\renewcommand{\arraystretch}{1.08}
\begin{adjustbox}{max width=\textwidth,center}
\begin{tabular}{l*{9}{c}}
\toprule
\multirow{2}{*}{\textbf{Method}} & \multicolumn{2}{c}{\textbf{Perception}} & \multicolumn{2}{c}{\textbf{OCR \& Chart}} & \multicolumn{2}{c}{\textbf{Symbolic}} & \multicolumn{2}{c}{\textbf{Multi-Image}} & \multirow{2}{*}{\textbf{Avg.}} \\
\cmidrule(lr){2-3}\cmidrule(lr){4-5}\cmidrule(lr){6-7}\cmidrule(lr){8-9}
& CountB. & PixMo & CC-OCR & ChartQA & VisPuz. & WeMath & MUIR & MMSI & \\

\midrule
\multicolumn{10}{c}{\textbf{\textit{Capability-Specific Experts}}} \\
\midrule
Perception-RL & 92.67 & 79.89 & 85.80 & 85.00 & 47.09 & 79.60 & 71.00 & 30.50 & 71.44 \\
\quad + Perception Skill & 93.69 & 82.35 & 85.82 & 85.08 & 47.00 & 79.54 & 71.12 & 30.60 & 71.90 \\[1pt]
OCR \& Chart-RL & 82.89 & 73.06 & 86.08 & 89.64 & 47.17 & 82.76 & 74.42 & 30.40 & 70.80 \\
\quad + OCR \& Chart Skill & 82.69 & 73.24 & 86.23 & 90.60 & 47.26 & 82.87 & 74.31 & 30.30 & 70.94 \\[1pt]
Symbolic-RL & 82.28 & 69.26 & 85.81 & 81.12 & 48.37 & 82.30 & 75.81 & 30.40 & 69.42 \\
\quad + Symbolic Skill & 82.08 & 69.07 & 85.78 & 81.36 & 48.80 & 84.20 & 75.85 & 30.50 & 69.71 \\[1pt]
Multi-Image-RL & 83.10 & 67.74 & 85.61 & 81.08 & 47.43 & 79.71 & 76.00 & 30.70 & 68.92 \\
\quad + Multi-Image Skill & 83.30 & 67.93 & 85.59 & 81.04 & 47.35 & 79.66 & 76.50 & 32.00 & 69.17 \\

\midrule
\multicolumn{10}{c}{\textbf{\textit{Unified-Model Baselines and Ablations}}} \\
\midrule
Zero-shot & 83.91 & 67.36 & 85.72 & 81.32 & 46.40 & 71.67 & 72.92 & 29.10 & 67.30 \\
SFT & 56.32 & 66.92 & 85.77 & 60.36 & 33.99 & 57.59 & 65.23 & 25.10 & 56.41 \\
Mix-RL & \underline{89.80} & \underline{73.80} & 86.01 & 85.00 & 40.10 & \textbf{82.90} & 67.80 & 29.10 & 69.31 \\
\addlinespace[2pt]
\rowcolor{cyan!3}
SCOPED (Parameter Merging) & 86.35 & 73.43 & 85.83 & 85.32 & \textbf{49.49} & 81.32 & 75.31 & 30.10 & 70.89 \\
\rowcolor{cyan!8}
SCOPED (w/o Skill Augmentation) & 89.41 & \textbf{79.89} & \textbf{88.89} & \underline{88.16} & 43.75 & 81.26 & \underline{76.27} & \underline{31.50} & \underline{72.39} \\
\rowcolor{cyan!15}
\textbf{SCOPED} & \textbf{90.43} & \textbf{79.89} & \underline{88.79} & \textbf{88.28} & \underline{46.58} & \underline{82.47} & \textbf{76.58} & \textbf{32.00} & \textbf{73.13} \\
\bottomrule
\end{tabular}
\end{adjustbox}

\vspace{-12pt}
\end{table}
As shown in Table~\ref{tab:ablation}, we conduct ablation studies to investigate the effectiveness of different multi-task training strategies and the individual components of \method{}. We first compare \method{} with SFT and Mix-RL to assess the benefits of our training paradigm. We then ablate key components of \method{} to examine their respective contributions. For reference, we also report the performance of capability-specific experts, each of which is optimized exclusively with task-specific rewards.

The evaluation results show that SFT underperforms the zero-shot backbone, suggesting that jointly fine-tuning on capability-specific data can introduce interference across capabilities. Although Mix-RL improves overall performance, it still underperforms the corresponding specialized experts on symbolic visual reasoning and multi-image reasoning, indicating that joint optimization may limit the acquisition of specialized capabilities. In contrast, \method{} achieves an average improvement of 3.82 points over Mix-RL, demonstrating the effectiveness of separating capability acquisition from capability integration. The skill-augmented expert results further show that matching each expert with its corresponding skill consistently improves the target capability, while its effects on non-target capabilities are limited and occasionally negative. This result highlights the capability-specific nature of the extracted guidance and motivates explicit skill-expert matching during distillation.

Furthermore, we compare two variants of \method{}, \method{} (Parameter Merging) and \method{} w/o Skill Augmentation, which use parameter merging and distillation without skill augmentation, respectively, to unify multiple capability-specific experts into a single model. \method{} (Parameter Merging) consistently underperforms the on-policy distillation variants, suggesting that directly merging expert parameters is insufficient to fully integrate complementary capabilities. In contrast, \method{} w/o Skill Augmentation already outperforms Mix-RL, indicating that token-level supervision from the corresponding experts along student-generated trajectories provides more fine-grained guidance for learning different capabilities. This allows the unified model to progressively absorb expertise that was independently acquired by the specialized experts. Building on this foundation, the full \method{} framework achieves a further average improvement, with gains across all four capability groups. These results indicate that explicit skills extracted from expert experience provide more targeted distillation signals, facilitating specialized capability distillation.

\subsection{Gradient Interference in Multi-Capability Training}
\label{sec:fusion_analysis}
\begin{figure}[t]
\centering
\captionsetup[subfigure]{font=footnotesize,skip=2pt}

\begin{subfigure}[t]{0.32\linewidth}
    \centering
    \includegraphics[width=\linewidth]{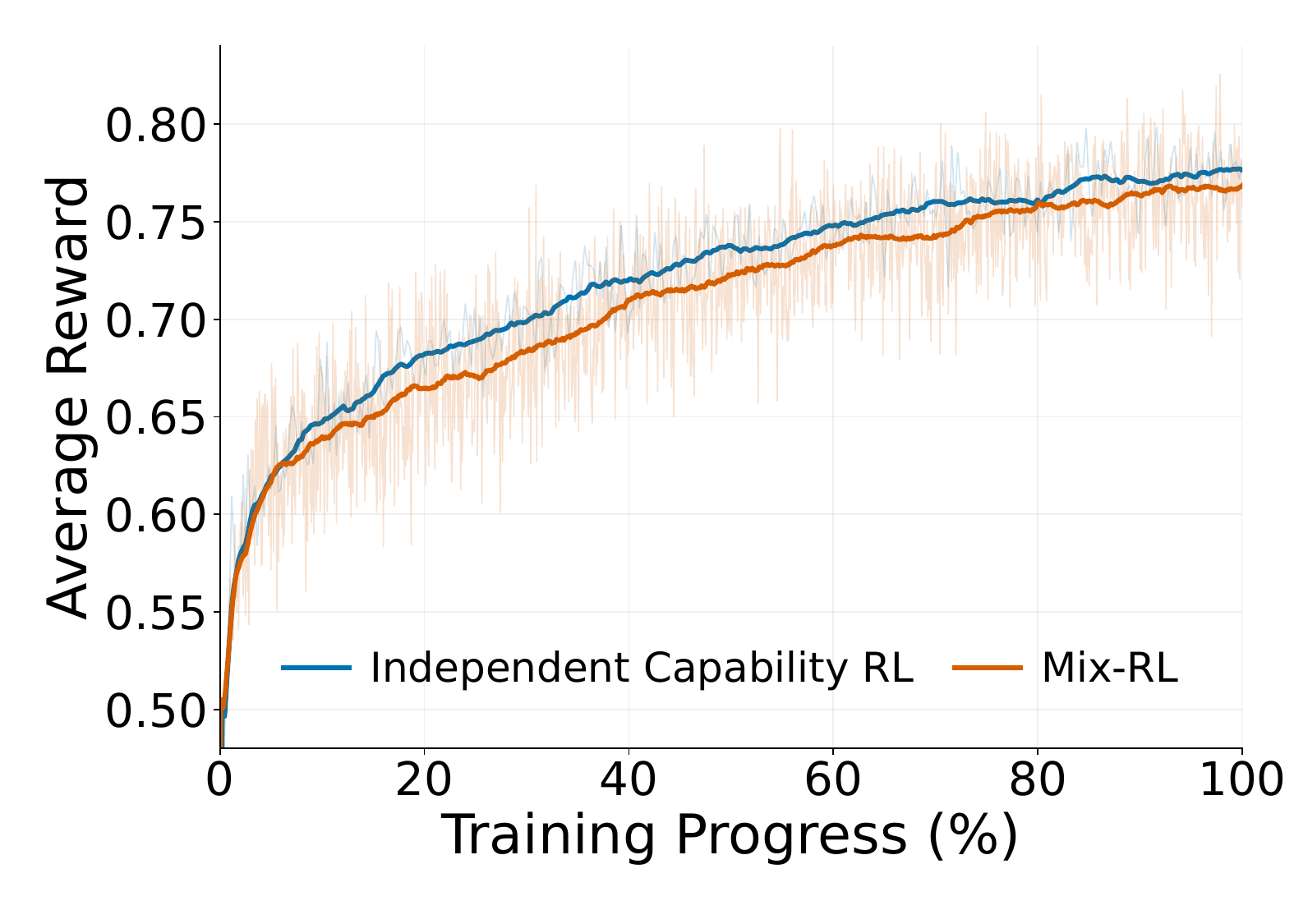}
    \caption{Training Reward.}
    \label{fig:reward_curve}
\end{subfigure}
\hfill
\begin{subfigure}[t]{0.32\linewidth}
    \centering
    \includegraphics[width=\linewidth]{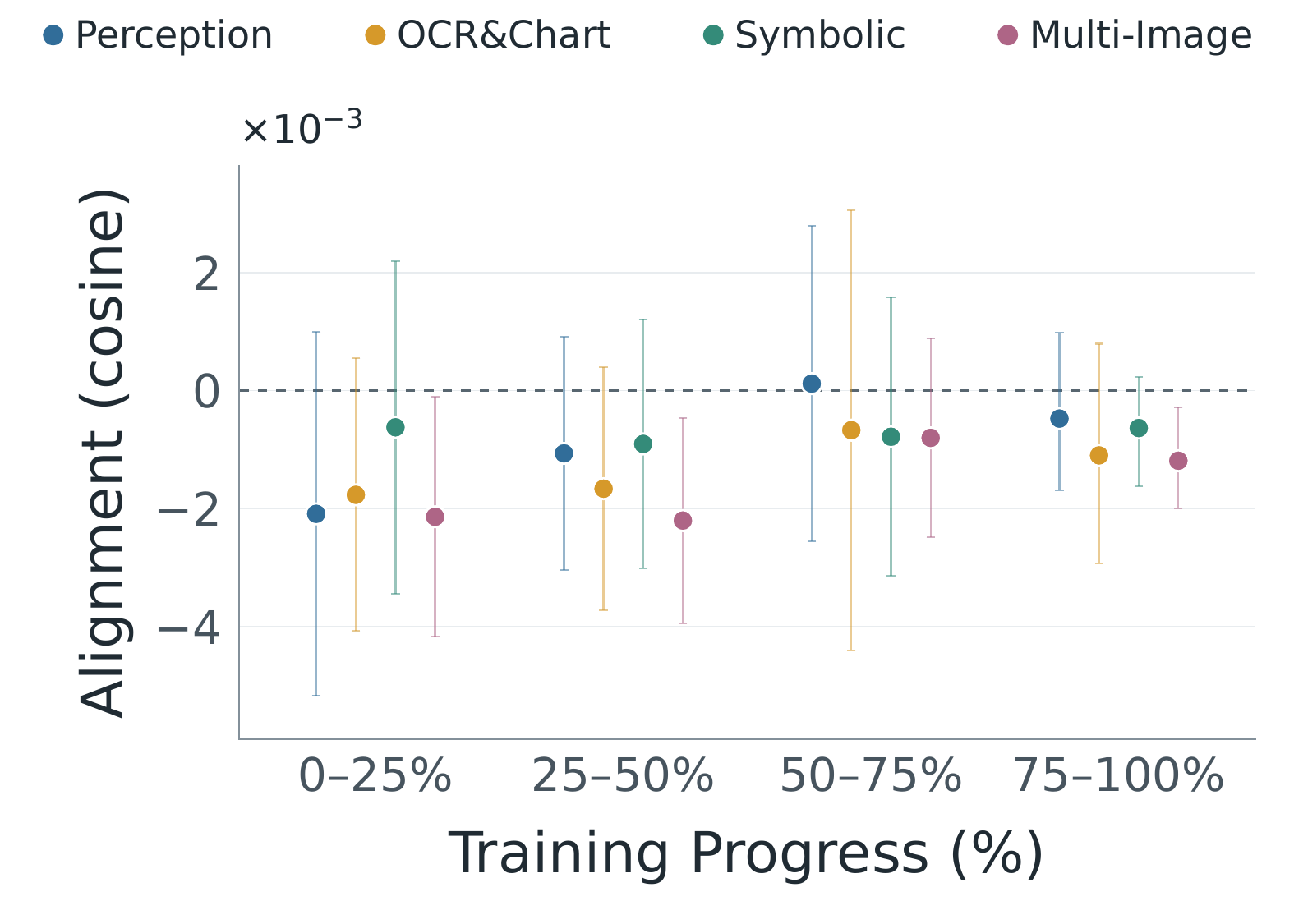}
    \caption{Capability-wise Update Alignment of the Mix-RL Model.}
    \label{fig:mixrl_alignment}
\end{subfigure}
\hfill
\begin{subfigure}[t]{0.32\linewidth}
    \centering
    \includegraphics[width=\linewidth]{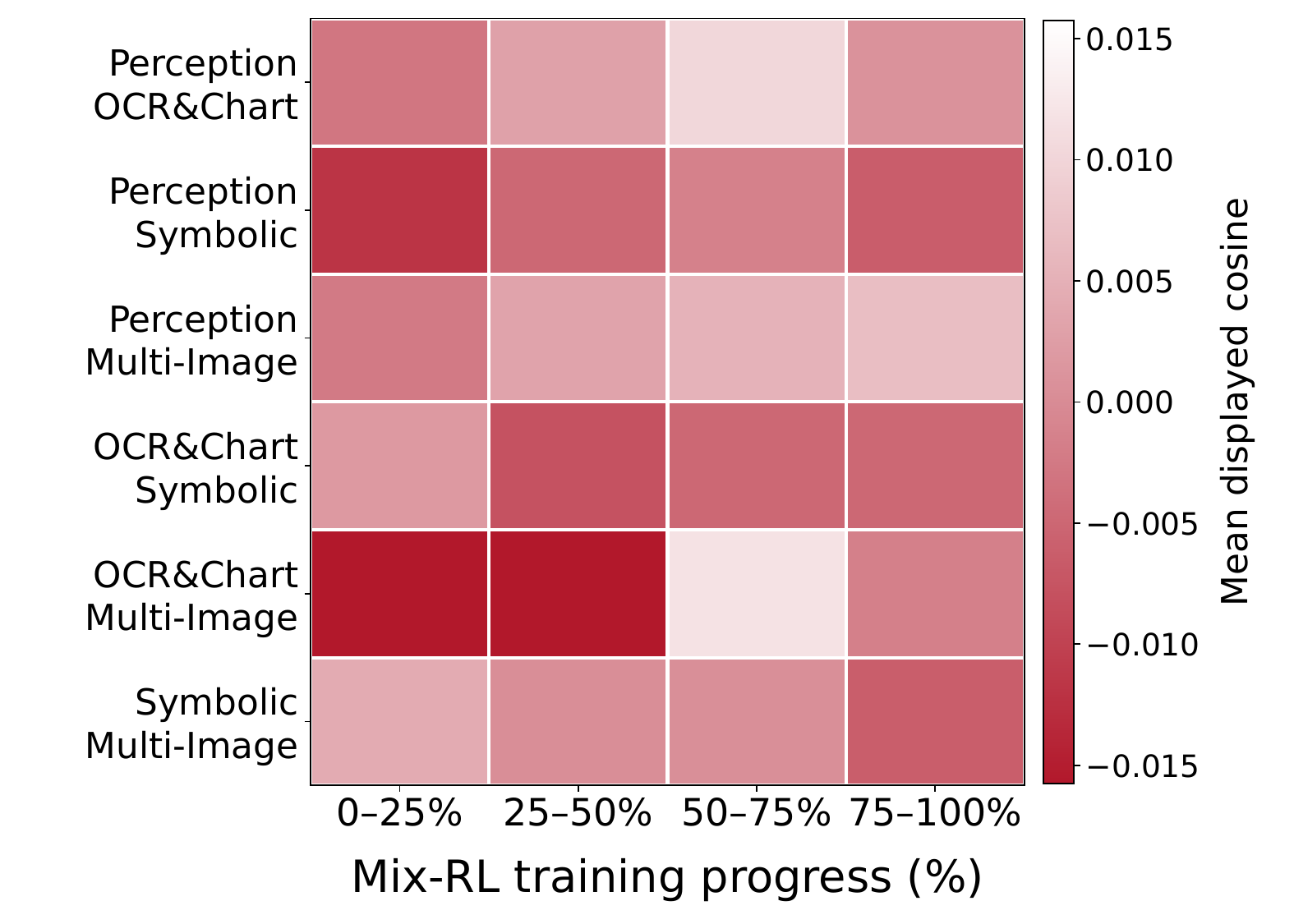}
    \caption{Pairwise Gradient Conflict.}
    \label{fig:pairwise_conflict}
\end{subfigure}

\caption{
Cross-Capability Interference in Mix-RL.
(a) Average training reward of independent capability RL and Mix-RL.
(b) Alignment of shared Mix-RL updates with capability-specific expert directions.
(c) Pairwise gradient cosine similarity across Mix-RL training stages.
}
\label{fig:gradient_conflict}
\vspace{-12pt}
\end{figure}

As illustrated in Figure~\ref{fig:gradient_conflict}, we investigate the cross-capability gradient interference phenomenon~\citep{yu2020gradient,liu2021conflict} during shared-policy optimization from three complementary perspectives: training reward, expert-guided update alignment, and pairwise gradient conflict.
We first evaluate the training dynamics in Figure~\ref{fig:reward_curve}. While both independent capability RL and Mix-RL steadily improve their average rewards, independent optimization maintains a consistent advantage throughout most of the training process. 
Although the performance gap narrows toward convergence, the overall trend clearly indicates that mixed-capability optimization structurally limits the gains achievable through capability-specific learning, preventing the model from fully benefiting from independently optimized capability objectives. To uncover the mechanistic cause of this limitation, we analyze how the realized Mix-RL updates align with the behaviors acquired by independently trained capability experts. We characterize this parameter-space update alignment using cosine similarity~\citep{xu2026thinking}. Specifically, for each capability, we derive a local expert-guided direction from the negative gradient of the KL divergence between the current Mix-RL policy and the corresponding fixed capability expert under matched response prefixes. We then compute its cosine similarity against the actual parameter displacement between adjacent Mix-RL checkpoints. In this formulation, positive values denote movement toward the expert, whereas near-zero or negative values indicate orthogonal or contradictory updates. As shown in Figure~\ref{fig:mixrl_alignment}, the shared updates show weak alignment with the expert-guided directions, with negative values observed across multiple capabilities and training stages. This confirms a severe see-saw effect: shared-policy optimization inevitably forces trade-offs, improving certain capabilities at the direct expense of others rather than consistently approximating all capability-specific experts simultaneously.

This interference is further supported by analyzing the capability-specific optimization directions themselves. At each Mix-RL checkpoint, we estimate capability-specific gradients using the corresponding capability data and compute pairwise cosine similarities among the four capabilities~\citep{yu2020gradient,liu2021conflict}. Positive cosine similarity indicates aligned optimization directions, whereas negative values indicate that two capabilities favor opposing local updates. Figure~\ref{fig:pairwise_conflict} reveals recurrent negative cosine similarities across multiple capability pairs at different training stages, providing direct empirical evidence that heterogeneous capabilities can favor competing local optimization directions. Together, these results demonstrate that shared-policy optimization is constrained by cross-capability interference. These results further support the motivation of \method{} to decouple capability acquisition into independently optimized experts prior to subsequent integration.
\subsection{Effectiveness of Skill-Augmented Distillation in \method{}}
\label{sec:skill_expert_retention}

\begin{figure}[t]
\centering
\captionsetup[subfigure]{font=footnotesize,skip=2pt}

\begin{subfigure}[t]{0.32\linewidth}
\centering
\includegraphics[width=\linewidth]{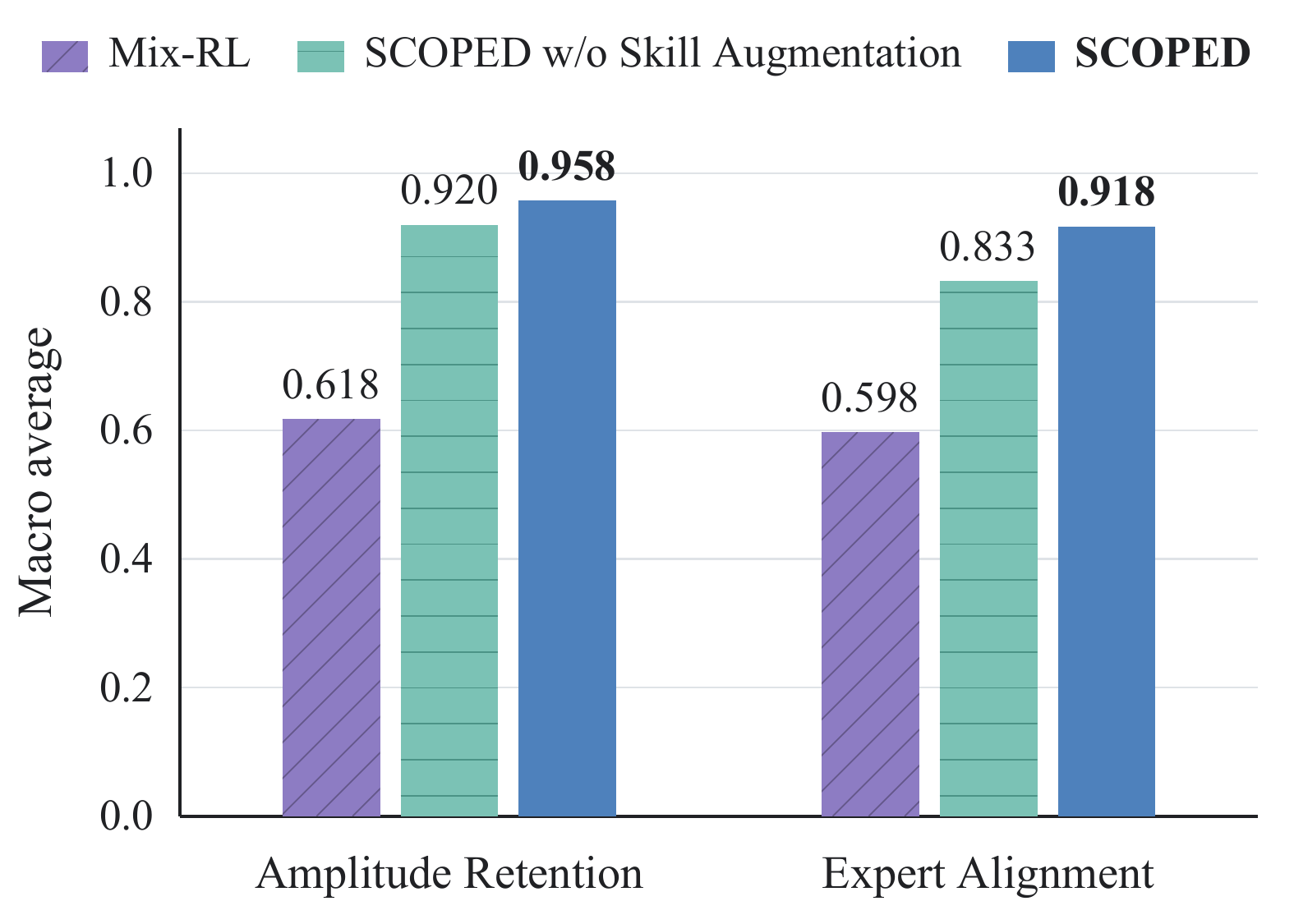}
\caption{Capability Retention of Experts.}
\label{fig:skill_retention_alignment}
\end{subfigure}
\hfill
\begin{subfigure}[t]{0.32\linewidth}
\centering
\includegraphics[width=\linewidth]{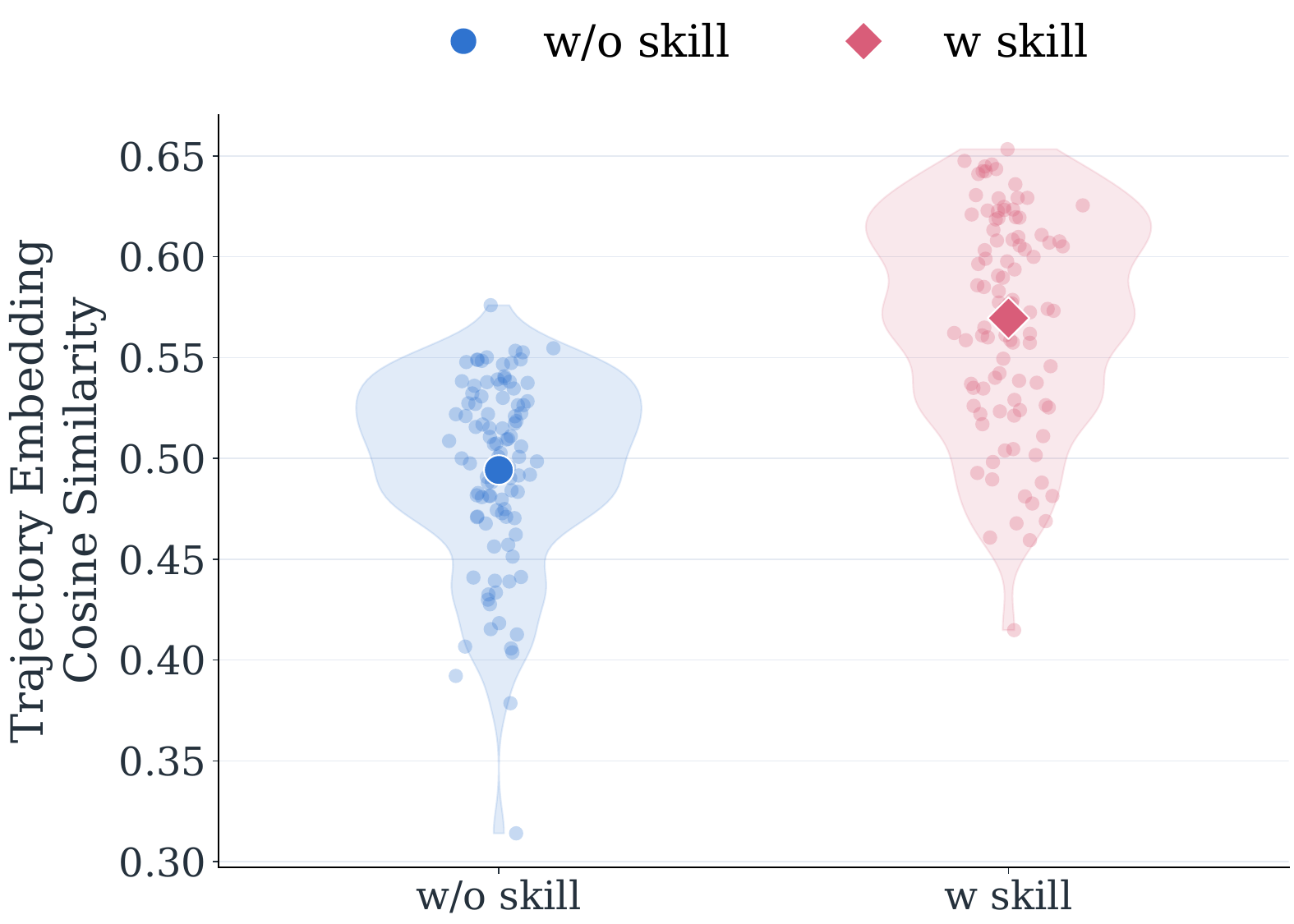}
\caption{Trajectory Consistency.}
\label{fig:trajectory_similarity}
\end{subfigure}
\hfill
\begin{subfigure}[t]{0.32\linewidth}
\centering
\includegraphics[width=\linewidth]{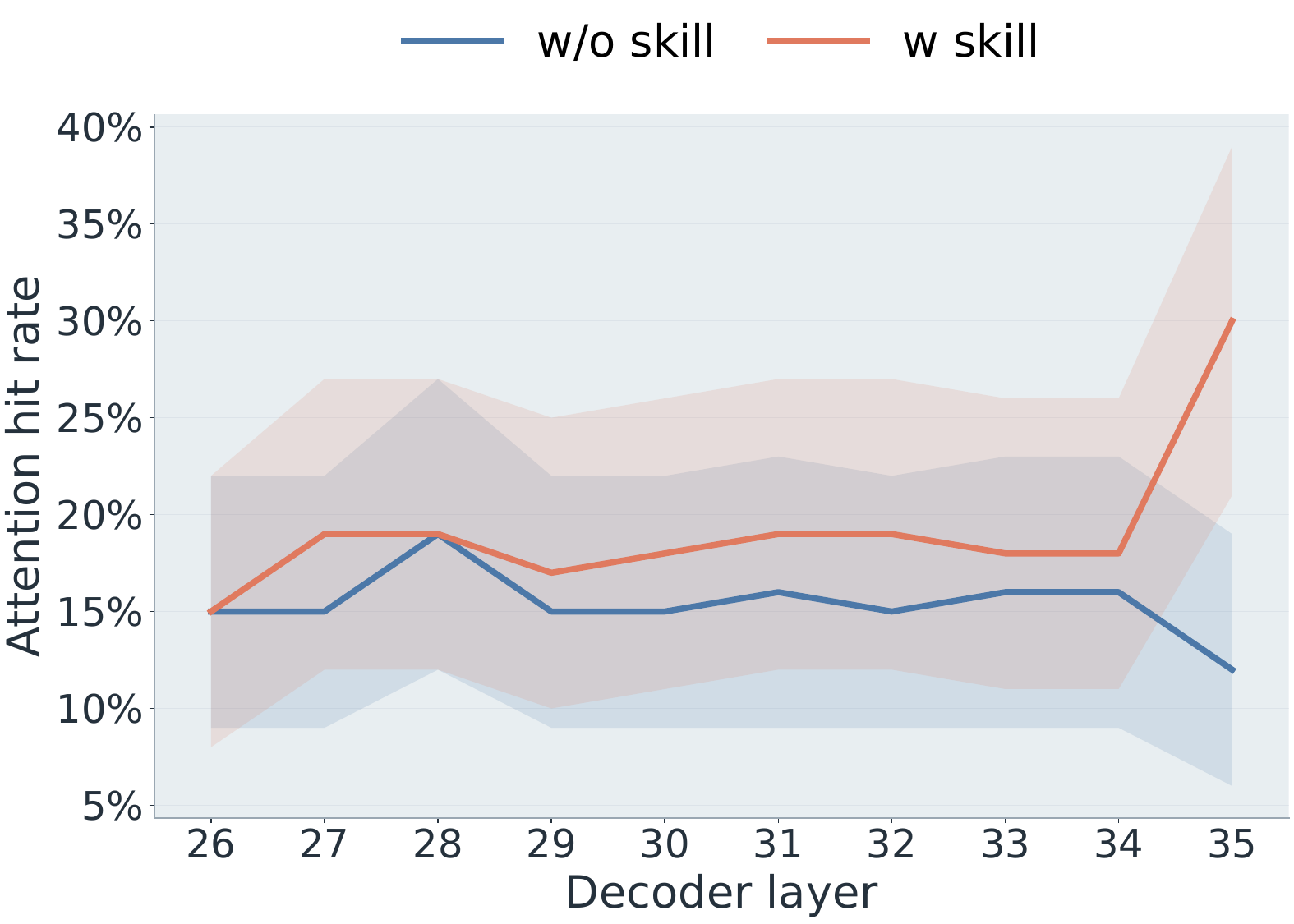}
\caption{Layer-wise Attention Hit Rate.}
\label{fig:attn_hit_layerwise}
\end{subfigure}
\caption{
Analysis of Skill-Augmented Expert Supervision.
(a) Capability retention and alignment, macro-averaged across four capabilities.
(b) Within-capability trajectory embedding similarity with and without skill conditioning.
(c) Layer-wise attention hit rate on task-relevant visual regions with and without skill conditioning.
}
\label{fig:skill_augmented_analysis}
\vspace{-12pt}
\end{figure}

In this section, we further investigate how skill augmentation regularizes expert supervision during the distillation process of \method{}. As illustrated in Figure~\ref{fig:skill_augmented_analysis}, we examine its effects from three complementary perspectives: the preservation of expert-induced behavior in the unified student, the consistency of expert trajectories across instances within each capability, and the utilization of task-relevant visual evidence.

We first characterize functional capability preservation using token-level log-probability shifts relative to the base model on fixed trajectories~\citep{klabunde2025similarity}. For each trajectory, we construct a shift vector from these log-probability changes and compare the unified student with the corresponding expert in terms of amplitude retention and expert alignment. Amplitude retention quantifies how much of the expert-induced shift is preserved by the student, while expert alignment measures the cosine similarity between their shift vectors. We evaluate models distilled from different capability-specific RL experts across visual understanding capabilities. As shown in Figure~\ref{fig:skill_retention_alignment}, \method{} achieves higher amplitude retention and expert alignment than Mix-RL, demonstrating better preservation of expert-induced functional behavior across capabilities. Notably, \method{} also outperforms \method{} w/o Skill Augmentation, indicating that the extracted skills help the student better acquire and reconcile complementary expert capabilities.

To further uncover the role of skill augmentation in \method{}, we evaluate the consistency of expert generation by encoding reasoning trajectories produced by each capability expert with and without its corresponding skill using Qwen3-VL-Embedding-2B~\citep{li2026qwen3}. Within each capability, we compute pairwise cosine similarities between trajectory embeddings generated from different inputs to obtain the distributions shown in Figure~\ref{fig:trajectory_similarity}. Higher similarity indicates more consistent expert behavior across instances. With skill conditioning, \method{} shifts the trajectory similarity distribution toward higher values, indicating that explicit skill guidance regularizes expert generation and promotes more consistent trajectory representations.
Finally, we investigate whether this trajectory-level regularization extends to visual evidence utilization. Building on findings that attention patterns in VLMs can effectively capture task-relevant visual grounding~\citep{kang2025your}, we compare each capability expert with and without its corresponding skill under identical inputs. At each decoder layer, we identify the most-attended visual token for each attention head and record a hit if it falls within the task-relevant region. We then average the resulting hit rates across attention heads and evaluation samples. As shown in Figure~\ref{fig:attn_hit_layerwise}, skill-augmented experts consistently achieve higher hit rates across most of the upper decoder layers, with the largest improvement observed in the final layer. This finding indicates that explicit skills steer capability experts toward more accurate allocation of attention to task-relevant visual evidence. Taken together, these multi-level analyses suggest that skill conditioning promotes more consistent expert trajectories and more effective visual grounding, while enabling the distilled student to better preserve expert-induced functional behavior.

\section{Conclusion}
This paper proposes \method{}, aiming at integrating complementary visual understanding capabilities into a unified Vision-Language Model through skill-decoupled on-policy expert distillation. \method{} independently optimizes capability-specific experts and abstracts reusable solution patterns from reward-evaluated trajectories into explicit capability skills. During on-policy distillation, these skills guide capability experts to provide structured supervision for the unified student. Our experiments demonstrate that \method{} consistently improves performance across diverse visual understanding capabilities. Our further analyses reveal cross-capability interference during mixed-capability optimization and show that skill augmentation improves the preservation and transfer of specialized expert capabilities.

\section*{AI Use Statement} Generative AI tools were used solely for language polishing and improving the clarity of the manuscript. All research ideas, experimental design, implementation, analysis, and conclusions were independently developed and verified by the authors, who take full responsibility for the content of the paper.

\bibliography{iclr2027_conference}

@article{bai2025qwen3,
  title={Qwen3-vl technical report},
  author={Bai, Shuai and Cai, Yuxuan and Chen, Ruizhe and Chen, Keqin and Chen, Xionghui and Cheng, Zesen and Deng, Lianghao and Ding, Wei and Gao, Chang and Ge, Chunjiang and others},
  journal={arXiv preprint arXiv:2511.21631},
  year={2025}
}

@article{wang2025internvl3,
  title={Internvl3. 5: Advancing open-source multimodal models in versatility, reasoning, and efficiency},
  author={Wang, Weiyun and Gao, Zhangwei and Gu, Lixin and Pu, Hengjun and Cui, Long and Wei, Xingguang and Liu, Zhaoyang and Jing, Linglin and Ye, Shenglong and Shao, Jie and others},
  journal={arXiv preprint arXiv:2508.18265},
  year={2025}
}

@article{ma2025one,
  title={One rl to see them all: Visual triple unified reinforcement learning},
  author={Ma, Yan and Du, Linge and Shen, Xuyang and Chen, Shaoxiang and Li, Pengfei and Ren, Qibing and Ma, Liz and Dai, Yuchao and Liu, Pengfei and Yan, Junjie},
  journal={arXiv preprint arXiv:2505.18129},
  year={2025}
}

@article{li2026qwen3,
  title={Qwen3-vl-embedding and qwen3-vl-reranker: A unified framework for state-of-the-art multimodal retrieval and ranking},
  author={Li, Mingxin and Zhang, Yanzhao and Long, Dingkun and Chen, Keqin and Song, Sibo and Bai, Shuai and Yang, Zhibo and Xie, Pengjun and Yang, An and Liu, Dayiheng and others},
  journal={arXiv preprint arXiv:2601.04720},
  year={2026}
}

@article{ni2026trace2skill,
  title={Trace2skill: Distill trajectory-local lessons into transferable agent skills},
  author={Ni, Jingwei and Liu, Yihao and Liu, Xinpeng and Sun, Yutao and Zhou, Mengyu and Cheng, Pengyu and Wang, Dexin and Zhao, Erchao and Jiang, Xiaoxi and Jiang, Guanjun},
  journal={arXiv preprint arXiv:2603.25158},
  year={2026}
}

@article{xia2026skillrl,
  title={Skillrl: Evolving agents via recursive skill-augmented reinforcement learning},
  author={Xia, Peng and Chen, Jianwen and Wang, Hanyang and Liu, Jiaqi and Zeng, Kaide and Wang, Yu and Han, Siwei and Zhou, Yiyang and Zhao, Xujiang and Chen, Haifeng and others},
  journal={arXiv preprint arXiv:2602.08234},
  year={2026}
}

@article{shi2026skill1,
  title={Skill1: Unified evolution of skill-augmented agents via reinforcement learning},
  author={Shi, Yaorui and Chen, Yuxin and Lu, Zhengxi and Miao, Yuchun and Liu, Shugui and Gu, Qi and Cai, Xunliang and Wang, Xiang and Zhang, An},
  journal={arXiv preprint arXiv:2605.06130},
  year={2026}
}

@article{yuan2026vision,
  title={Vision-opd: Learning to see fine details for multimodal llms via on-policy self-distillation},
  author={Yuan, Qianhao and Lou, Jie and Yu, Xing and Lin, Hongyu and Sun, Le and Han, Xianpei and Lu, Yaojie},
  journal={arXiv preprint arXiv:2605.18740},
  year={2026}
}

@article{xiao2026mimo,
  title={Mimo-v2-flash technical report},
  author={Xiao, Bangjun and Xia, Bingquan and Yang, Bo and Gao, Bofei and Shen, Bowen and Zhang, Chen and He, Chenhong and Lou, Chiheng and Luo, Fuli and Wang, Gang and others},
  journal={arXiv preprint arXiv:2601.02780},
  year={2026}
}

@article{xu2026deepseek,
  title={Deepseek-v4: Towards highly efficient million-token context intelligence},
  author={Xu, Anyi and Lin, Bangcai and Xue, Bing and Wang, Bingxuan and Xu, Bingzheng and Wu, Bochao and Zhang, Bowei and Lin, Chaofan and Dong, Chen and Ling, Chenchen and others},
  journal={arXiv preprint arXiv:2606.19348},
  year={2026}
}

@article{ma2026mopd,
  title={MOPD: Multi-Teacher On-Policy Distillation for Capability Integration in LLM Post-Training},
  author={Ma, Wenhan and Wei, Jianyu and Zhao, Liang and Zhang, Hailin and Xiao, Bangjun and Li, Lei and Yang, Qibin and Gao, Bofei and Wang, Yudong and Li, Rang and others},
  journal={arXiv preprint arXiv:2606.30406},
  year={2026}
}

@article{wang2026skill,
  title={Skill-sd: Skill-conditioned self-distillation for multi-turn llm agents},
  author={Wang, Hao and Wang, Guozhi and Xiao, Han and Zhou, Yufeng and Pan, Yue and Wang, Jichao and Xu, Ke and Wen, Yafei and Ruan, Xiaohu and Chen, Xiaoxin and others},
  journal={arXiv preprint arXiv:2604.10674},
  year={2026}
}

@article{ye2026policy,
  title={On-policy context distillation for language models},
  author={Ye, Tianzhu and Dong, Li and Wu, Xun and Huang, Shaohan and Wei, Furu},
  journal={arXiv preprint arXiv:2602.12275},
  year={2026}
}

@article{zhao2026self,
  title={Self-Distilled Reasoner: On-Policy Self-Distillation for Large Language Models},
  author={Zhao, Siyan and Xie, Zhihui and Liu, Mengchen and Huang, Jing and Pang, Guan and Chen, Feiyu and Grover, Aditya},
  journal={arXiv preprint arXiv:2601.18734},
  year={2026}
}

@inproceedings{agarwal2024policy,
  title={On-policy distillation of language models: Learning from self-generated mistakes},
  author={Agarwal, Rishabh and Vieillard, Nino and Zhou, Yongchao and Stanczyk, Piotr and Ramos Garea, Sabela and Geist, Matthieu and Bachem, Olivier},
  booktitle={International Conference on Learning Representations},
  volume={2024},
  pages={21246--21263},
  year={2024}
}

@article{sarch2026vero,
  title={Vero: An open RL recipe for general visual reasoning},
  author={Sarch, Gabriel and Cai, Linrong and Wang, Qunzhong and Wu, Haoyang and Chen, Danqi and Liu, Zhuang},
  journal={arXiv preprint arXiv:2604.04917},
  year={2026}
}

@inproceedings{masry2022chartqa,
  title={Chartqa: A benchmark for question answering about charts with visual and logical reasoning},
  author={Masry, Ahmed and Tan, Jia Qing and Joty, Shafiq and Hoque, Enamul and others},
  booktitle={Findings of the association for computational linguistics: ACL 2022},
  pages={2263--2279},
  year={2022}
}

@article{song2025visualpuzzles,
  title={Visualpuzzles: Decoupling multimodal reasoning evaluation from domain knowledge},
  author={Song, Yueqi and Ou, Tianyue and Kong, Yibo and Li, Zecheng and Neubig, Graham and Yue, Xiang},
  journal={arXiv preprint arXiv:2504.10342},
  year={2025}
}

@inproceedings{qiao2025we,
  title={We-math: Does your large multimodal model achieve human-like mathematical reasoning?},
  author={Qiao, Runqi and Tan, Qiuna and Dong, Guanting and MinhuiWu, MinhuiWu and Sun, Chong and Song, Xiaoshuai and Wang, Jiapeng and Gongque, Zhuoma and Lei, Shanglin and Zhang, Yifan and others},
  booktitle={Proceedings of the 63rd Annual Meeting of the Association for Computational Linguistics (Volume 1: Long Papers)},
  pages={20023--20070},
  year={2025}
}

@inproceedings{wang2025muirbench,
  title={Muirbench: A comprehensive benchmark for robust multi-image understanding},
  author={Wang, Fei and Fu, Xingyu and Huang, James Y and Li, Zekun and Liu, Qin and Liu, Xiaogeng and Ma, Mingyu Derek and Xu, Nan and Zhou, Wenxuan and Zhang, Kai and others},
  booktitle={International Conference on Learning Representations},
  volume={2025},
  pages={62624--62650},
  year={2025}
}

@article{yang2025mmsi,
  title={Mmsi-bench: A benchmark for multi-image spatial intelligence},
  author={Yang, Sihan and Xu, Runsen and Xie, Yiman and Yang, Sizhe and Li, Mo and Lin, Jingli and Zhu, Chenming and Chen, Xiaochen and Duan, Haodong and Yue, Xiangyu and others},
  journal={arXiv preprint arXiv:2505.23764},
  year={2025}
}

@inproceedings{paiss2023teaching,
  title={Teaching clip to count to ten},
  author={Paiss, Roni and Ephrat, Ariel and Tov, Omer and Zada, Shiran and Mosseri, Inbar and Irani, Michal and Dekel, Tali},
  booktitle={Proceedings of the IEEE/CVF international conference on computer vision},
  pages={3170--3180},
  year={2023}
}

@inproceedings{deitke2025molmo,
  title={Molmo and pixmo: Open weights and open data for state-of-the-art vision-language models},
  author={Deitke, Matt and Clark, Christopher and Lee, Sangho and Tripathi, Rohun and Yang, Yue and Park, Jae Sung and Salehi, Mohammadreza and Muennighoff, Niklas and Lo, Kyle and Soldaini, Luca and others},
  booktitle={Proceedings of the Computer Vision and Pattern Recognition Conference},
  pages={91--104},
  year={2025}
}

@article{hong2025glm,
  title={Glm-4.5 v and glm-4.1 v-thinking: Towards versatile multimodal reasoning with scalable reinforcement learning},
  author={Hong, Wenyi and Yu, Wenmeng and Gu, Xiaotao and Wang, Guo and Gan, Guobing and Tang, Haomiao and Cheng, Jiale and Qi, Ji and Ji, Junhui and Pan, Lihang and others},
  journal={arXiv preprint arXiv:2507.01006},
  year={2025}
}

@article{bai2025qwen2,
  author = {Bai, Shuai and Chen, Keqin and Liu, Xuejing and Wang, Jialin and Ge, Wenbin and Song, Sibo and Dang, Kai and Wang, Peng and Wang, Shijie and Tang, Jun and others},
  title = {{Qwen2.5-VL} Technical Report},
  journal = {ArXiv preprint},
  year = {2025},
  url = {https://arxiv.org/abs/2502.13923}
}

@article{xiaomi2025mimovl,
  title={{MiMo-VL Technical Report}},
  author={{Xiaomi LLM-Core Team}},
  journal={arXiv preprint arXiv:2506.03569},
  year={2025}
}

@article{yu2026dapo,
  title={Dapo: An open-source llm reinforcement learning system at scale},
  author={Yu, Qiying and Zhang, Zheng and Zhu, Ruofei and Yuan, Yufeng and Zuo, Xiaochen and Yue, Yu and Dai, Weinan and Fan, Tiantian and Liu, Gaohong and Liu, Lingjun and others},
  journal={Advances in Neural Information Processing Systems},
  volume={38},
  pages={113222--113244},
  year={2026}
}

@article{xu2026cc,
  title={CC-OCR V2: Benchmarking Large Multimodal Models for Literacy in Real-world Document Processing},
  author={Xu, Zhipeng and Ji, Junhao and Chen, Zulong and Liu, Zhenghao and Liu, Qing and Peng, Chunyi and Qin, Zubao and Xu, Ze and Wan, Jianqiang and Tang, Jun and others},
  journal={arXiv preprint arXiv:2605.03903},
  year={2026}
}

@misc{openai2025gpt52,
  author       = {{OpenAI}},
  title        = {Introducing {GPT-5.2}},
  year         = {2025},
  month        = dec,
  howpublished = {\url{https://openai.com/index/introducing-gpt-5-2/}},
  note         = {Published December 11, 2025}
}

@misc{openai2026gpt54,
  author       = {{OpenAI}},
  title        = {Introducing {GPT-5.4}},
  year         = {2026},
  month        = mar,
  howpublished = {\url{https://openai.com/index/introducing-gpt-5-4/}},
  note         = {Published March 5, 2026}
}

@misc{anthropic2026sonnet46,
  author       = {{Anthropic}},
  title        = {Introducing {Claude Sonnet 4.6}},
  year         = {2026},
  month        = feb,
  howpublished = {\url{https://www.anthropic.com/news/claude-sonnet-4-6}},
  note         = {Published February 17, 2026}
}

@misc{anthropic2026opus46,
  author       = {{Anthropic}},
  title        = {Introducing {Claude Opus 4.6}},
  year         = {2026},
  month        = feb,
  howpublished = {\url{https://www.anthropic.com/news/claude-opus-4-6}},
  note         = {Published February 5, 2026}
}

@article{sheng2024hybridflow,
  title   = {HybridFlow: A Flexible and Efficient RLHF Framework},
  author  = {Guangming Sheng and Chi Zhang and Zilingfeng Ye and Xibin Wu and Wang Zhang and Ru Zhang and Yanghua Peng and Haibin Lin and Chuan Wu},
  year    = {2024},
  journal = {arXiv preprint arXiv: 2409.19256}
}

@misc{zheng2025easyr1,
  author = {Yaowei Zheng and Junting Lu and Shenzhi Wang and Zhangchi Feng and Dongdong Kuang and Yuwen Xiong},
  title = {{EasyR1}: An Efficient, Scalable, Multi-Modality {RL} Training Framework},
  year = {2025},
  howpublished = {\url{https://github.com/hiyouga/EasyR1}},
  note = {GitHub repository}
}

@inproceedings{gu2024minillm,
  title={Minillm: Knowledge distillation of large language models},
  author={Gu, Yuxian and Dong, Li and Wei, Furu and Huang, Minlie},
  booktitle={The twelfth international conference on learning representations},
  year={2024}
}

@article{jiang2024mantis,
  title={Mantis: Interleaved multi-image instruction tuning},
  author={Jiang, Dongfu and He, Xuan and Zeng, Huaye and Wei, Cong and Ku, Max and Liu, Qian and Chen, Wenhu},
  journal={arXiv preprint arXiv:2405.01483},
  year={2024}
}

@article{li2024llava,
  title={Llava-next-interleave: Tackling multi-image, video, and 3d in large multimodal models},
  author={Li, Feng and Zhang, Renrui and Zhang, Hao and Zhang, Yuanhan and Li, Bo and Li, Wei and Ma, Zejun and Li, Chunyuan},
  journal={arXiv preprint arXiv:2407.07895},
  year={2024}
}

@inproceedings{xiong2026lang2act,
  title={Lang2act: Fine-grained visual reasoning through self-emergent linguistic toolchains},
  author={Xiong, Yuqi and Peng, Chunyi and Xu, Zhipeng and Liu, Zhenghao and Chen, Zulong and Yan, Yukun and Wang, Shuo and Gu, Yu and Yu, Ge},
  booktitle={Findings of the Association for Computational Linguistics: ACL 2026},
  pages={8375--8399},
  year={2026}
}

@article{sun2025visrag,
  title={Visrag 2.0: Evidence-guided multi-image reasoning in visual retrieval-augmented generation},
  author={Sun, Yubo and Peng, Chunyi and Yan, Yukun and Yu, Shi and Liu, Zhenghao and Chen, Chi and Liu, Zhiyuan and Sun, Maosong},
  journal={arXiv preprint arXiv:2510.09733},
  year={2025}
}

@article{peng2025mixture,
  title={Mixture-of-Retrieval Experts for Reasoning-Guided Multimodal Knowledge Exploitation},
  author={Peng, Chunyi and Xu, Zhipeng and Liu, Zhenghao and Li, Yishan and Yan, Yukun and Wang, Shuo and Gu, Yu and Yu, Minghe and Yu, Ge and Sun, Maosong},
  journal={arXiv preprint arXiv:2505.22095},
  year={2025}
}

@inproceedings{liu2026visionreasoner,
  title={Visionreasoner: Unified reasoning-integrated visual perception via reinforcement learning},
  author={Liu, Yuqi and Qu, Tianyuan and Zhong, Zhisheng and Peng, Bohao and Liu, Shu and Yu, Bei and Jia, Jiaya},
  booktitle={International Conference on Learning Representations},
  volume={2026},
  pages={94069--94086},
  year={2026}
}

@article{xia2025visionary,
  title={Visionary-r1: Mitigating shortcuts in visual reasoning with reinforcement learning},
  author={Xia, Jiaer and Zang, Yuhang and Gao, Peng and Li, Sharon and Zhou, Kaiyang},
  journal={arXiv preprint arXiv:2505.14677},
  year={2025}
}

@inproceedings{zhang2026improving,
  title={Improving the reasoning of multi-image grounding in mllms via reinforcement learning},
  author={Zhang, Bob and Li, Haoran and Zhang, Tao and Li, Jianan and Yan, Cilin and Liu, Xikai and Cai, Jiayin and Hao, Yanbin},
  booktitle={ICASSP 2026-2026 IEEE International Conference on Acoustics, Speech and Signal Processing (ICASSP)},
  pages={12667--12671},
  year={2026},
  organization={IEEE}
}

@inproceedings{wei2026optmerge,
  title={Optmerge: Unifying multimodal llm capabilities and modalities via model merging},
  author={Wei, Yongxian and Cheng, Runxi and Jin, Weike and Yang, Enneng and Shen, Li and Hou, Lu and Du, Sinan and Yuan, Chun and Cao, Xiaochun and Tao, Dacheng},
  booktitle={International Conference on Learning Representations},
  volume={2026},
  pages={61571--61595},
  year={2026}
}

@article{yu2020gradient,
  title={Gradient surgery for multi-task learning},
  author={Yu, Tianhe and Kumar, Saurabh and Gupta, Abhishek and Levine, Sergey and Hausman, Karol and Finn, Chelsea},
  journal={Advances in neural information processing systems},
  volume={33},
  pages={5824--5836},
  year={2020}
}

@article{liu2021conflict,
  title={Conflict-averse gradient descent for multi-task learning},
  author={Liu, Bo and Liu, Xingchao and Jin, Xiaojie and Stone, Peter and Liu, Qiang},
  journal={Advances in neural information processing systems},
  volume={34},
  pages={18878--18890},
  year={2021}
}

@article{jiang2026co,
  title={Co-reinforcement learning for unified multimodal understanding and generation},
  author={Jiang, Jingjing and Si, Chongjie and Luo, Jun and Zhang, Hanwang and Ma, Chao},
  journal={Advances in Neural Information Processing Systems},
  volume={38},
  pages={5823--5846},
  year={2026}
}

@article{cai2026advancing,
  title={Advancing general-purpose reasoning models with modular gradient surgery},
  author={Cai, Min and Liang, Yu and Wang, Longzheng and Wang, Yan and Zhang, Yueyang and Xia, Long and Sun, Zhiyuan and Ye, Xi and Shi, Daiting},
  journal={arXiv preprint arXiv:2602.02301},
  year={2026}
}

@article{chen2024odin,
  title={Odin: Disentangled reward mitigates hacking in rlhf},
  author={Chen, Lichang and Zhu, Chen and Soselia, Davit and Chen, Jiuhai and Zhou, Tianyi and Goldstein, Tom and Huang, Heng and Shoeybi, Mohammad and Catanzaro, Bryan},
  journal={arXiv preprint arXiv:2402.07319},
  year={2024}
}

@article{helff2026llms,
  title={LLMs gaming verifiers: RLVR can lead to reward hacking},
  author={Helff, Lukas and Delfosse, Quentin and Steinmann, David and H{\"a}rle, Ruben and Shindo, Hikaru and Schramowski, Patrick and Stammer, Wolfgang and Kersting, Kristian and Friedrich, Felix},
  journal={arXiv preprint arXiv:2604.15149},
  year={2026}
}

@article{guo2025deepseek,
  title={Deepseek-r1: Incentivizing reasoning capability in llms via reinforcement learning},
  author={Guo, Daya and Yang, Dejian and Zhang, Haowei and Song, Junxiao and Wang, Peiyi and Zhu, Qihao and Xu, Runxin and Zhang, Ruoyu and Ma, Shirong and Bi, Xiao and others},
  journal={arXiv preprint arXiv:2501.12948},
  year={2025}
}

@inproceedings{gao2023scaling,
  title={Scaling laws for reward model overoptimization},
  author={Gao, Leo and Schulman, John and Hilton, Jacob},
  booktitle={International conference on machine learning},
  pages={10835--10866},
  year={2023},
  organization={PMLR}
}

@inproceedings{liu2026mitigating,
  title={Mitigating Judgment Preference Bias in Large Language Models through Group-Based Polling},
  author={Liu, Shuliang and Xu, Zhipeng and Liu, Zhenghao and Yan, Yukun and Yu, Minghe and Gu, Yu and Chen, Chong and Xie, Huiyuan and Yu, Ge},
  booktitle={Findings of the Association for Computational Linguistics: ACL 2026},
  pages={1448--1464},
  year={2026}
}

@article{huang2026skill,
  title={Skill-Conditioned Gated Self-Distillation for LLM Reasoning},
  author={Huang, Jiazhen and Chen, Xiao and Luo, Xiao and Dai, Yong and Hu, Senkang and Zhao, Yuzhi},
  journal={arXiv preprint arXiv:2605.28791},
  year={2026}
}

@article{klabunde2025similarity,
  title={Similarity of neural network models: A survey of functional and representational measures},
  author={Klabunde, Max and Schumacher, Tobias and Strohmaier, Markus and Lemmerich, Florian},
  journal={ACM Computing Surveys},
  volume={57},
  number={9},
  pages={1--52},
  year={2025},
  publisher={ACM New York, NY}
}

@inproceedings{kang2025your,
  title={Your large vision-language model only needs a few attention heads for visual grounding},
  author={Kang, Seil and Kim, Jinyeong and Kim, Junhyeok and Hwang, Seong Jae},
  booktitle={2025 IEEE/CVF Conference on Computer Vision and Pattern Recognition (CVPR)},
  pages={9339--9350},
  year={2025},
  organization={IEEE}
}

@inproceedings{xu2026thinking,
  title={Thinking-free policy initialization makes distilled reasoning models more effective and efficient reasoners},
  author={Xu, Xin and Bai, Clive and Yang, Kai and Chen, Tianhao and Wang, Yang and Yang, Saiyong and Yang, Can},
  booktitle={International Conference on Learning Representations},
  volume={2026},
  pages={24062--24086},
  year={2026}
}

@article{peng2026memory,
  title={Memory Shot for Long-Term Dialogue},
  author={Peng, Chunyi and Xin, Haidong and Sheng, Xuanshuo and Dai, Xin and Liu, Zhenghao and Wang, Shuo and Yan, Yukun and Chen, Zulong and Gu, Yu and Yu, Ge},
  journal={arXiv preprint arXiv:2606.28338},
  year={2026}
}

@inproceedings{xin2026metamem,
  title={Metamem: Evolving meta-memory for knowledge utilization through self-reflective symbolic optimization},
  author={Xin, Haidong and Li, Xinze and Liu, Zhenghao and Yan, Yukun and Wang, Shuo and Yang, Cheng and Gu, Yu and Yu, Ge and Sun, Maosong},
  booktitle={Findings of the Association for Computational Linguistics: ACL 2026},
  pages={5473--5492},
  year={2026}
}

@article{gao2026teaching,
  title={Teaching llms to learn tool trialing and execution through environment interaction},
  author={Gao, Xingjie and Huang, Pengcheng and Liu, Zhenghao and Yan, Yukun and Wang, Shuo and Chen, Zulong and Qian, Chen and Yu, Ge and Gu, Yu},
  journal={arXiv preprint arXiv:2601.12762},
  year={2026}
}
\bibliographystyle{iclr2027_conference}
\clearpage
\appendix

\section{APPENDIX}
\subsection{Dataset Details}
\label{app:data_statistics}

We evaluate models on eight benchmarks spanning four complementary capability groups. Fine-grained visual perception is assessed using CountBenchQA~\citep{paiss2023teaching} and PixMo-Count~\citep{deitke2025molmo}. OCR and chart understanding are evaluated on the recognition track of CC-OCR V2~\citep{xu2026cc} and ChartQA~\citep{masry2022chartqa}. Symbolic visual reasoning is evaluated on VisualPuzzles~\citep{song2025visualpuzzles} and We-Math~\citep{qiao2025we}. Multi-image reasoning is evaluated on MUIRBench~\citep{wang2025muirbench} and MMSI-Bench~\citep{yang2025mmsi}. Table~\ref{tab:dataset_statistics} summarizes the evaluation scale and primary capability of each benchmark. The reported sample counts correspond to the instances retained by our evaluation pipeline after preprocessing.

\begin{table}[H]
\centering

\caption{
Statistics of the Benchmarks.
The number of evaluation samples corresponds to the instances retained
by our evaluation pipeline.
}
\label{tab:dataset_statistics}

\small
\setlength{\tabcolsep}{5pt}
\renewcommand{\arraystretch}{1.08}

\begin{adjustbox}{max width=\textwidth,center}
\begin{tabular}{llrl}
\toprule
\textbf{Capability}
& \textbf{Benchmark}
& \textbf{\#Eval.}
& \textbf{Description} \\
\midrule

Perception
& CountBenchQA
& 491
& Object counting in natural images \\

Perception
& PixMo-Count
& 527
& Fine-grained counting in diverse natural scenes \\

\midrule

OCR \& Chart
& CC-OCR V2
& 1,790
& Multilingual and natural-scene text recognition \\

OCR \& Chart
& ChartQA
& 2,500
& Visual and logical reasoning over charts \\

\midrule

Symbolic
& VisualPuzzles
& 1,168
& General visual and abstract puzzle reasoning \\

Symbolic
& We-Math
& 1,740
& Hierarchical visual mathematical reasoning \\

\midrule

Multi-Image
& MUIRBench
& 2,600
& Robust reasoning over multiple related images \\

Multi-Image
& MMSI-Bench
& 1,000
& Multi-image spatial intelligence and reasoning \\

\bottomrule
\end{tabular}
\end{adjustbox}
\end{table}

For capability-specific expert training, we construct the reinforcement learning corpus from six public multimodal data sources and organize the retained samples into four capability-specific subsets. The fine-grained perception and grounding capability uses PixMo-Count together with capability-aligned samples from Orsta-Data-47K~\citep{ma2025one}. The OCR, chart, and document understanding capability uses SlideVQA together with corresponding samples from Orsta-Data-47K. The symbolic and mathematical reasoning capability uses We-Math 2.0 Pro together with corresponding samples from Orsta-Data-47K. The multi-image spatial reasoning capability uses Mantis-Instruct~\citep{jiang2024mantis} and M4-Instruct~\citep{li2024llava}. After preprocessing and filtering, the resulting training corpus contains 37,999 samples. Table~\ref{tab:training_data_statistics} reports the data sources and sample counts for each capability.

    \begin{table}[H]
    \centering 
    \caption{Statistics of the capability-specific reinforcement learning data. The reported counts 
    correspond to the samples retained after preprocessing and filtering.}
    \label{tab:training_data_statistics}
    \small
    \setlength{\tabcolsep}{5pt}
    \renewcommand{\arraystretch}{1.08}
    \begin{tabular}{l l r}
    \toprule
    \textbf{Capability} & \textbf{Data Sources} & \textbf{\# Samples} \\
    \midrule
    Perception \& Grounding & PixMo-Count, Orsta-Data-47K & 9,499 \\
    OCR \& Chart & SlideVQA, Orsta-Data-47K & 9,500 \\
    Symbolic Reasoning & We-Math 2.0 Pro, Orsta-Data-47K & 9,500 \\
    Multi-Image Reasoning & Mantis-Instruct, M4-Instruct & 9,500 \\
    \midrule
    \textbf{Total} & -- & \textbf{37,999} \\
    \bottomrule
    \end{tabular}
    \end{table}

\subsection{Additional Implementation Details}
\label{app:implementation_details}

For the main experiments, all capability-specialized experts and the unified student are initialized from Qwen3-VL-8B-Instruct~\citep{bai2025qwen3}, while the additional model-scale experiments use Qwen3-VL-4B-Instruct under the same experimental setup. Capability-specialized expert training is implemented with EasyR1~\citep{zheng2025easyr1}, while skill-decoupled on-policy expert distillation is implemented with veRL~\citep{sheng2024hybridflow}. All experiments are conducted on NVIDIA A800 80GB GPUs. Unless otherwise specified, the four capability experts share the same optimization and rollout configurations, differing only in their capability-specific training data and reward functions. During skill-decoupled on-policy expert distillation, the four independently optimized capability experts provide supervision to the unified student, with each expert conditioned on its corresponding capability skill while the student receives only the original multimodal input. The main hyperparameters for both capability-specialized expert training and skill-decoupled on-policy expert distillation are summarized in Table~\ref{tab:training_hyperparameters}.

\begin{table}[t]
\centering
\caption{Hyperparameters for capability-specialized expert training and skill-decoupled on-policy expert distillation.}
\label{tab:training_hyperparameters}

\scriptsize
\setlength{\tabcolsep}{3pt}
\renewcommand{\arraystretch}{1.08}

\begin{minipage}[t]{0.48\linewidth}
\centering
\textbf{(a) Capability-Specialized Expert Training}

\vspace{3pt}
\begin{tabular}{@{}lr@{}}
\toprule
\textbf{Hyperparameter} & \textbf{Value} \\
\midrule
Training epochs & 3 \\
Rollout batch size & 128 \\
Actor global batch size & 64 \\
Responses per prompt & 8 \\
Maximum prompt length & 8,192 \\
Maximum response length & 8,192 \\
Learning rate & $1\times10^{-6}$ \\
Weight decay & $1\times10^{-2}$ \\
Warmup ratio & 0.0 \\
Maximum gradient norm & 1.0 \\
KL coefficient & $1\times10^{-2}$ \\
Rollout temperature & 1.0 \\
Top-$p$ & 1.0 \\
Tensor parallel size & 2 \\
Minimum image pixels & 262,144 \\
Maximum image pixels & 524,288 \\
Training precision & bfloat16 \\
\bottomrule
\end{tabular}
\end{minipage}
\hfill
\begin{minipage}[t]{0.48\linewidth}
\centering
\textbf{(b) Skill-Decoupled On-Policy Expert Distillation}

\vspace{3pt}
\begin{tabular}{@{}lr@{}}
\toprule
\textbf{Hyperparameter} & \textbf{Value} \\
\midrule
Training epochs & 3 \\
Rollout batch size & 256 \\
Actor mini-batch size & 64 \\
Responses per prompt & 1 \\
Maximum prompt length & 12,288 \\
Maximum response length & 8,096 \\
Learning rate & $1\times10^{-6}$ \\
Weight decay & $1\times10^{-2}$ \\
Warmup ratio & 0.0 \\
Maximum gradient norm & 1.0 \\
KL penalty & Disabled \\
Rollout temperature & 1.0 \\
Top-$p$ & 1.0 \\
Student tensor parallel size & 2 \\
Teacher tensor parallel size & 1 \\
Distillation loss mode & \texttt{k\_1} \\
Training precision & bfloat16 \\
\bottomrule
\end{tabular}
\end{minipage}

\end{table}

\subsection{Reward Design}
\label{app:reward_design}

We design task-specific automatic rewards to accommodate the heterogeneous output spaces of the training tasks. For textual predictions, the answer is extracted from the \texttt{<answer>} field when available and normalized by lowercasing, removing punctuation and articles, and collapsing consecutive spaces. Structured predictions are additionally checked against their required output formats.

\paragraph{Fine-Grained Perception and Visual Grounding.}
For perception tasks with textual outputs, we use token-level F1 between the normalized prediction and reference answer. For visual grounding, the prediction is parsed as a list of labeled bounding boxes. Predicted and ground-truth boxes with the same normalized label are greedily matched according to their intersection-over-union (IoU). The grounding reward is defined as
\begin{equation}
r_{\mathrm{ground}}=\lambda_{\mathrm{fmt}}r_{\mathrm{fmt}}+\lambda_{0.5}F_{1,\tau=0.5}+\lambda_{0.75}F_{1,\tau=0.75}+\lambda_{\mathrm{IoU}}r_{\mathrm{mIoU}},
\label{eq:grounding_reward}
\end{equation}
where \(r_{\mathrm{fmt}}\) indicates whether the predicted boxes can be successfully parsed, \(F_{1,\tau=0.5}\) and \(F_{1,\tau=0.75}\) denote label-aware detection F1 at IoU thresholds of \(0.5\) and \(0.75\), respectively, and \(r_{\mathrm{mIoU}}\) is the mean IoU obtained from one-to-one box matching. We set \((\lambda_{\mathrm{fmt}},\lambda_{0.5},\lambda_{0.75},\lambda_{\mathrm{IoU}})=(0.1,0.4,0.3,0.2)\).

\paragraph{OCR and Chart Understanding.}
For OCR and chart-understanding tasks, we adopt token-level F1 as the
training reward:
\begin{equation}
r_{\mathrm{text}}=\frac{2PR}{P+R},
\label{eq:text_f1_reward}
\end{equation}
where \(P\) and \(R\) are computed from the token overlap between the
normalized prediction and reference answer. Boolean outputs are mapped to
a unified \textit{yes}/\textit{no} representation before matching. Empty
predictions or predictions without overlapping tokens receive zero reward.

\paragraph{Multimodal Symbolic Reasoning.}
For symbolic reasoning tasks, we employ format-gated answer accuracy:
\begin{equation}
r_{\mathrm{symbolic}}=\mathbb{I}_{\mathrm{fmt}}\,\operatorname{Acc}(a,y),
\label{eq:symbolic_reward}
\end{equation}
where \(\mathbb{I}_{\mathrm{fmt}}\) indicates whether the response contains
a valid \texttt{<answer>...</answer>} field. Numerical answers are compared
by exact numerical equality after removing commas and optional percentage
symbols. Textual answers are normalized before matching, and a prediction
is considered correct if it matches any valid reference answer. Responses
that do not satisfy the required answer format receive zero reward.

\paragraph{Multi-Image Spatial Reasoning.}
Multi-image reasoning tasks also use format-gated accuracy:
\begin{equation}
r_{\mathrm{multi}}=\mathbb{I}_{\mathrm{fmt}}\,\operatorname{Acc}_{\mathrm{type}}(a,y),
\label{eq:multiimage_reward}
\end{equation}
where the matching function is selected according to the answer type.
Multiple-choice questions require an exact option-letter match, while
image-selection questions compare the referenced image indices. Yes/no
expressions are mapped to canonical labels, and spatial descriptions are
mapped to normalized relations such as \textit{left}, \textit{above},
\textit{inside}, and \textit{not touching}. Remaining free-form answers
are evaluated using normalized containment matching. As in symbolic
reasoning, malformed responses receive zero overall reward.
\subsection{Additional Ablation on Model Scale}
\label{app:model_scale_ablation}
\begin{table}[t]
\centering
\caption{Ablation Study. All models are built on Qwen3-VL-4B. We compare training and capability integration strategies, with capability-specific experts shown as references for assessing expertise retention. The best and second-best results are highlighted in \textbf{bold} and \underline{underlined}, respectively.}
\label{tab:ablation_4b}
\small
\setlength{\tabcolsep}{2.8pt}
\renewcommand{\arraystretch}{1.08}
\begin{adjustbox}{max width=\textwidth,center}
\begin{tabular}{l*{9}{c}}
\toprule
\multirow{2}{*}{\textbf{Method}} & \multicolumn{2}{c}{\textbf{Perception}} & \multicolumn{2}{c}{\textbf{OCR \& Chart}} & \multicolumn{2}{c}{\textbf{Symbolic}} & \multicolumn{2}{c}{\textbf{Multi-Image}} & \multirow{2}{*}{\textbf{Avg.}} \\
\cmidrule(lr){2-3}\cmidrule(lr){4-5}\cmidrule(lr){6-7}\cmidrule(lr){8-9}
& CountB. & PixMo & CC-OCR & ChartQA & VisPuz. & WeMath & MUIR & MMSI & \\

\midrule
\multicolumn{10}{c}{\textbf{\textit{Capability-Specific Experts}}} \\
\midrule
Perception-RL & 88.39 & 73.24 & 84.76 & 87.64 & 40.41 & 77.87 & 73.54 & 30.20 & 69.51 \\
\quad + Perception Skill & 89.41 & 75.33 & 84.79 & 87.72 & 40.33 & 77.82 & 73.65 & 30.30 & 69.92 \\[1pt]
OCR \& Chart-RL & 86.35 & 68.12 & 85.25 & 90.36 & 42.29 & 80.06 & 74.00 & 29.20 & 69.45 \\
\quad + OCR \& Chart Skill & 86.15 & 68.31 & 85.46 & 91.28 & 42.38 & 80.17 & 73.88 & 29.10 & 69.59 \\[1pt]
Symbolic-RL & 85.34 & 64.33 & 83.60 & 85.00 & 35.62 & 78.91 & 73.81 & 29.50 & 67.01 \\
\quad + Symbolic Skill & 85.13 & 64.14 & 83.57 & 85.24 & 36.04 & 80.34 & 73.85 & 29.60 & 67.24 \\[1pt]
Multi-Image-RL & 79.63 & 59.96 & 84.25 & 86.64 & 41.44 & 75.29 & 75.65 & 30.60 & 66.68 \\
\quad + Multi-Image Skill & 79.84 & 60.15 & 84.22 & 86.60 & 41.35 & 75.23 & 76.15 & 31.50 & 66.88 \\

\midrule
\multicolumn{10}{c}{\textbf{\textit{Unified-Model Baselines and Ablations}}} \\
\midrule
Zero-shot & 81.47 & 64.90 & 82.86 & 85.60 & 35.10 & 76.26 & 72.12 & \textbf{30.20} & 66.06 \\
SFT & 63.54 & 44.40 & 74.67 & 65.36 & 31.93 & 46.32 & 61.08 & 24.80 & 51.51 \\
Mix-RL & 86.35 & 70.97 & 83.10 & 87.60 & 34.76 & \textbf{80.29} & 69.92 & 29.10 & 67.76 \\
\addlinespace[2pt]
\rowcolor{cyan!3}
SCOPED (Parameter Merging) & 86.35 & 71.16 & 84.49 & 86.60 & \underline{37.93} & 76.95 & 73.88 & 26.60 & 68.00 \\
\rowcolor{cyan!8}
SCOPED (w/o Skill Augmentation) & \underline{87.98} & \underline{73.81} & \underline{84.74} & \underline{88.96} & 37.50 & 78.16 & \underline{75.69} & 29.20 & \underline{69.51} \\
\rowcolor{cyan!15}
\textbf{SCOPED} & \textbf{88.39} & \textbf{74.38} & \textbf{85.03} & \textbf{89.48} & \textbf{38.01} & \underline{78.85} & \textbf{76.15} & \underline{29.40} & \textbf{69.96} \\
\bottomrule
\end{tabular}
\end{adjustbox}

\vspace{2pt}
\end{table}
We further conduct the same ablation study on Qwen3-VL-4B to examine whether the effectiveness of our design remains consistent at a smaller model scale. As shown in Table~\ref{tab:ablation_4b}, SFT substantially underperforms the zero-shot backbone, while Mix-RL yields only a modest overall improvement and remains below the corresponding capability-specific experts on several target capabilities. In contrast, independently trained experts achieve stronger performance on their respective capabilities, showing that the benefit of separating capability acquisition from integration remains evident with a smaller backbone.

The skill-augmented expert results exhibit a trend consistent with Qwen3-VL-8B. Conditioning each expert on its matched skill improves performance on the corresponding target capability, while the effects on non-target capabilities remain limited or mixed. This further supports the capability-specific nature of the extracted guidance and the use of explicit skill-expert matching during distillation.

For capability integration, we compare \method{} (Parameter Merging), \method{} w/o Skill Augmentation, and the full \method{} framework using the same capability-specific experts. \method{} w/o Skill Augmentation outperforms parameter merging, showing that on-policy expert supervision provides a more effective mechanism for transferring independently acquired capabilities than direct parameter-space merging. Incorporating skill augmentation further improves the unified model, yielding the best average performance among the compared integration variants. Together with the Qwen3-VL-8B results, these consistent trends indicate that the effectiveness of \method{} is not restricted to a particular backbone scale.

\subsection{Skill Construction Details}
\label{skill_construct}

We provide additional implementation details for the capability-wise skill abstraction procedure introduced in Sec.~\ref{sec:skill_mopd}. Skill construction uses only training data and training-set rollouts, while the held-out evaluation set is used exclusively for final evaluation. For each capability, a single skill is constructed from the trajectories generated by its corresponding optimized expert.

\paragraph{Trajectory Curation.}
For each training instance, the optimized capability expert samples multiple rollout trajectories, each of which is evaluated by the corresponding capability-specific reward function. To reduce redundancy, we deduplicate repeated trajectories within each task subtype and balance the retained trajectories across subtypes so that high-resource tasks do not dominate the subsequent skill construction process.

\paragraph{Reward-Based Trajectory Grouping.}
Following Sec.~\ref{sec:skill_mopd}, the curated trajectories are divided into successful and failed cohorts according to their capability-specific rewards. This grouping provides complementary evidence about effective solution patterns and recurring failure modes.

\paragraph{Trajectory-Level Reflection.}
The optimized capability expert performs trajectory-level reflection over the successful and failed cohorts using outcome-specific reflection prompts. Successful trajectories are used to identify reusable solution strategies and verification behaviors, while failed trajectories are analyzed to diagnose recurring failure mechanisms and derive corrective guidance. The resulting reflections are aggregated into a capability-specific evidence ledger containing both effective behaviors and failure-avoidance guidance.

\paragraph{Capability Skill Synthesis.}
Based on the aggregated trajectory-derived evidence, the same capability expert synthesizes a single capability-level skill containing reusable procedural guidance for the corresponding capability. The resulting skill is subsequently used to condition the expert during skill-decoupled on-policy expert distillation. This procedure produces one skill for each of the four capabilities. Detailed reflection and skill-induction prompts are provided in Appendix~\ref{app:prompt_examples}.

\paragraph{Ablation of Trajectory Sources.}
Table~\ref{tab:trajectory_source_ablation} examines the contribution of successful and failed trajectories to skill construction. Using either successful or failed trajectories alone improves the corresponding capability experts, while combining both sources achieves the strongest overall performance. This indicates that successful solution patterns and corrective failure experience provide complementary information for capability skill abstraction.
\begin{table*}[t]
\centering
\caption{
Ablation of trajectory sources used for capability skill abstraction on Qwen3-VL-8B.
The baseline reports the absolute performance of the corresponding capability-specialized experts, while the remaining rows report performance changes relative to the baseline.
}
\label{tab:trajectory_source_ablation}

\small
\setlength{\tabcolsep}{3.6pt}
\renewcommand{\arraystretch}{1.08}

\begin{adjustbox}{max width=\textwidth,center}
\begin{tabular}{l*{9}{c}}
\toprule

\multirow{2}{*}{\textbf{Condition}}
& \multicolumn{2}{c}{\textbf{Perception}}
& \multicolumn{2}{c}{\textbf{OCR \& Chart}}
& \multicolumn{2}{c}{\textbf{Symbolic}}
& \multicolumn{2}{c}{\textbf{Multi-Image}}
& \multirow{2}{*}{\textbf{Avg.}} \\

\cmidrule(lr){2-3}
\cmidrule(lr){4-5}
\cmidrule(lr){6-7}
\cmidrule(lr){8-9}

& \textbf{CountB.}
& \textbf{PixMo}
& \textbf{CC-OCR}
& \textbf{ChartQA}
& \textbf{VisPuz.}
& \textbf{WeMath}
& \textbf{MUIR}
& \textbf{MMSI}
& \\

\midrule
\multicolumn{10}{l}{\textit{Baseline}} \\
Capability-RL Expert
& 92.67
& 79.89
& 86.08
& 89.64
& 48.37
& 82.30
& 76.00
& 30.70
& 73.21 \\

\midrule
\multicolumn{10}{l}{\textit{Skill Abstraction (delta from Capability-RL Expert)}} \\

\hspace{2mm}+ Failure
& +0.41
& +1.52
& +0.07
& +0.56
& +0.17
& +0.75
& +0.35
& +0.60
& +0.55 \\

\hspace{2mm}+ Success
& +0.61
& +1.14
& +0.09
& +0.48
& +0.26
& +1.15
& +0.19
& +0.80
& +0.59 \\

\hspace{2mm}+ Combined
& \textbf{+1.02}
& \textbf{+2.47}
& \textbf{+0.15}
& \textbf{+0.96}
& \textbf{+0.43}
& \textbf{+1.90}
& \textbf{+0.50}
& \textbf{+1.30}
& \textbf{+1.09} \\

\bottomrule
\end{tabular}
\end{adjustbox}
\end{table*}

\subsection{Details of Cross-Capability Gradient Analysis}
\label{app:gradient_conflict}

We provide the computation protocol for the two gradient-based analyses in Figure~\ref{fig:gradient_conflict}. Figure~\ref{fig:mixrl_alignment} examines whether the realized Mix-RL update follows the directions induced by independently trained capability experts, while Figure~\ref{fig:pairwise_conflict} directly measures pairwise conflicts among capability-specific local optimization directions under the same Mix-RL parameters.

\paragraph{Expert-Guided Update Alignment.}
For each capability $c$, we fix the independently trained capability expert $\pi_{\theta_c}$ and select a fixed set of 32 samples. At a Mix-RL checkpoint $\theta_t$, both the current Mix-RL policy and the corresponding expert are evaluated under identical response prefixes. Let $q_{c,n,s}$ and $p_{\theta_t,c,n,s}$ denote the expert and Mix-RL predictive distributions, respectively, at token position $s$. We define the expert-matching objective as
\begin{equation}
L_c^{\mathrm{exp}}(\theta_t)=\frac{1}{N_c}\sum_{n=1}^{N_c}\frac{1}{T_{c,n}}\sum_{s=1}^{T_{c,n}}D_{\mathrm{KL}}\!\left(q_{c,n,s}\,\|\,p_{\theta_t,c,n,s}\right),
\label{eq:expert_guided_kl}
\end{equation}
where $N_c=32$ and $T_{c,n}$ denotes the number of valid response positions for sample $n$. The corresponding expert-guided direction is defined as the negative gradient
\begin{equation}
\mathbf{g}_c^{\mathrm{exp}}(\theta_t)=-\nabla_{\theta}L_c^{\mathrm{exp}}(\theta)\big|_{\theta=\theta_t}.
\label{eq:expert_guided_direction}
\end{equation}
Given two adjacent Mix-RL checkpoints $t<t'$, the realized parameter displacement is
\begin{equation}
\Delta\theta_{t\rightarrow t'}=\theta_{t'}-\theta_t.
\label{eq:mixrl_parameter_displacement}
\end{equation}
We measure whether this realized update follows the corresponding expert-guided direction using
\begin{equation}
A_c(t\rightarrow t')=\frac{\left\langle\Delta\theta_{t\rightarrow t'},\mathbf{g}_c^{\mathrm{exp}}(\theta_t)\right\rangle}{\left\|\Delta\theta_{t\rightarrow t'}\right\|_2\left\|\mathbf{g}_c^{\mathrm{exp}}(\theta_t)\right\|_2}.
\label{eq:expert_guided_alignment}
\end{equation}
All four capabilities are compared against the same Mix-RL parameter displacement within each checkpoint interval. Positive values indicate that the realized update moves toward the corresponding expert behavior, whereas negative values indicate movement against the expert-guided direction. For Figure~\ref{fig:mixrl_alignment}, checkpoint-level alignments are grouped into four training stages and summarized within each capability.

\paragraph{Reward-Based Capability Optimization Direction.}
To directly analyze interactions among capability objectives, we separately estimate a reward-based local optimization direction for each capability under the same Mix-RL parameters. At checkpoint $\theta_t$, we sample $K$ responses $\{y_{c,n}^{(k)}\}_{k=1}^{K}$ for each input $x_{c,n}$ from capability $c$ and obtain their rewards $\{r_{c,n}^{(k)}\}_{k=1}^{K}$. The group-normalized advantage is
\begin{equation}
A_{c,n}^{(k)}=\frac{r_{c,n}^{(k)}-\bar r_{c,n}}{\sigma_{c,n}+\epsilon},
\label{eq:gradient_advantage}
\end{equation}
where $\bar r_{c,n}$ and $\sigma_{c,n}$ are the mean and population standard deviation of the $K$ rewards, respectively, and $\epsilon=10^{-6}$. Groups with $\sigma_{c,n}<\epsilon$ are discarded. Let $\mathcal{V}_{c,t}$ denote the remaining valid responses and $N_{c,t}$ their total token count. The resulting local optimization direction is
\begin{equation}
\mathbf{u}_c(\theta_t)=\frac{1}{N_{c,t}}\sum_{(n,k)\in\mathcal{V}_{c,t}}A_{c,n}^{(k)}\sum_{s=1}^{|y_{c,n}^{(k)}|}\left.\nabla_{\theta}\log\pi_{\theta}\!\left(y_{c,n,s}^{(k)}\mid x_{c,n},y_{c,n,<s}^{(k)}\right)\right|_{\theta=\theta_t}.
\label{eq:capability_local_direction}
\end{equation}
Gradients are computed over the language-policy parameters, with the visual parameters frozen.

\paragraph{Pairwise Cross-Capability Conflict.}
At each Mix-RL checkpoint, all capability-specific directions are estimated under the same shared parameters $\theta_{\mathrm{mix},t}$. For each unordered capability pair $(c,c')$, we compute
\begin{equation}
C_{c,c'}(t)=\frac{\left\langle\mathbf{u}_c(\theta_{\mathrm{mix},t}),\mathbf{u}_{c'}(\theta_{\mathrm{mix},t})\right\rangle}{\left\|\mathbf{u}_c(\theta_{\mathrm{mix},t})\right\|_2\left\|\mathbf{u}_{c'}(\theta_{\mathrm{mix},t})\right\|_2}.
\label{eq:cross_capability_cosine}
\end{equation}
With four capabilities, this yields six unordered capability pairs at each checkpoint. Negative cosine similarity indicates locally opposing optimization directions and therefore provides direct evidence of cross-capability gradient conflict. For Figure~\ref{fig:pairwise_conflict}, the checkpoint-level cosine similarities are grouped into four training stages, and the mean similarity of each capability pair within each stage is reported.

\subsection{Cross-Capability Gradient Interference in Mix-RL}
\label{sec:fusion_analysis-cross}

We further investigate cross-capability interference by comparing the parameter-update alignment of Mix-RL against that of the independently trained capability experts $\{\pi_{\theta_c}\}_{c=1}^C$. Specifically, we evaluate capability-specific optimization directions $\mathbf{u}_c(\cdot)$ estimated directly from the RL rewards $r_c$.

For Mix-RL, let $\theta_{\mathrm{mix},t}$ denote the shared policy parameters at training checkpoint $t$, and $\Delta\theta_{\mathrm{mix},t}=\theta_{\mathrm{mix},t+1}-\theta_{\mathrm{mix},t}$ denote the realized parameter displacement between adjacent checkpoints. At each checkpoint $t$, we estimate the optimization directions for all $C$ capabilities under the shared parameters $\theta_{\mathrm{mix},t}$. The alignment between the shared update and each capability-specific direction is defined as
\begin{equation}
M_c(t)=\frac{\langle\Delta\theta_{\mathrm{mix},t},\mathbf{u}_c(\theta_{\mathrm{mix},t})\rangle}{\|\Delta\theta_{\mathrm{mix},t}\|_2\|\mathbf{u}_c(\theta_{\mathrm{mix},t})\|_2}.
\label{eq:mixrl_update_alignment}
\end{equation}

Similarly, during the independent RL training of each capability expert $\pi_{\theta_c}$, let $\theta_{c,t}$ denote its intermediate parameters at checkpoint $t$. We define the corresponding parameter displacement as $\Delta\theta_{c,t}=\theta_{c,t+1}-\theta_{c,t}$ and measure its alignment with its respective optimization direction:
\begin{equation}
S_c(t)=\frac{\langle\Delta\theta_{c,t},\mathbf{u}_c(\theta_{c,t})\rangle}{\|\Delta\theta_{c,t}\|_2\|\mathbf{u}_c(\theta_{c,t})\|_2}.
\label{eq:independent_update_alignment}
\end{equation}

Here, positive alignment indicates that the realized update follows the corresponding capability-specific optimization direction, whereas negative alignment reflects an opposing update. To further characterize cross-capability interference, we compute pairwise cosine similarities among the capability-specific directions $\{\mathbf{u}_c(\theta_{\mathrm{mix},t})\}_{c=1}^C$ at each Mix-RL checkpoint, where negative values indicate conflicting local optimization directions.

\begin{figure}[t]
    \centering
    \captionsetup[subfigure]{font=footnotesize,skip=2pt}
    \begin{subfigure}[t]{0.495\linewidth}
        \centering
        \includegraphics[width=\linewidth,trim=12 8 12 8,clip]{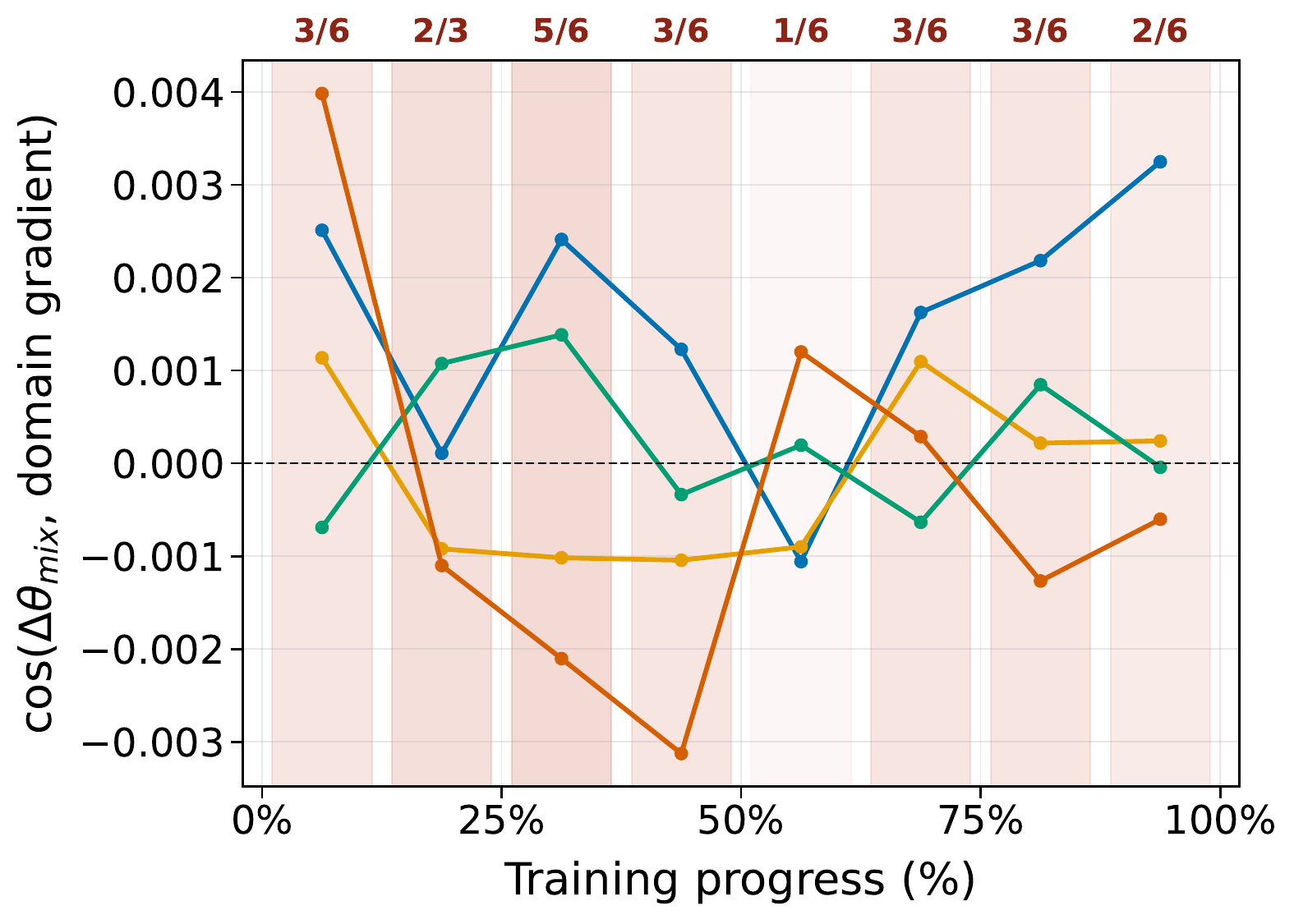}
        \caption{Mix-RL Update Alignment}
        \label{fig:mixrl_reward_alignment}
    \end{subfigure}\hfill
    \begin{subfigure}[t]{0.495\linewidth}
        \centering
        \includegraphics[width=\linewidth,trim=12 8 12 8,clip]{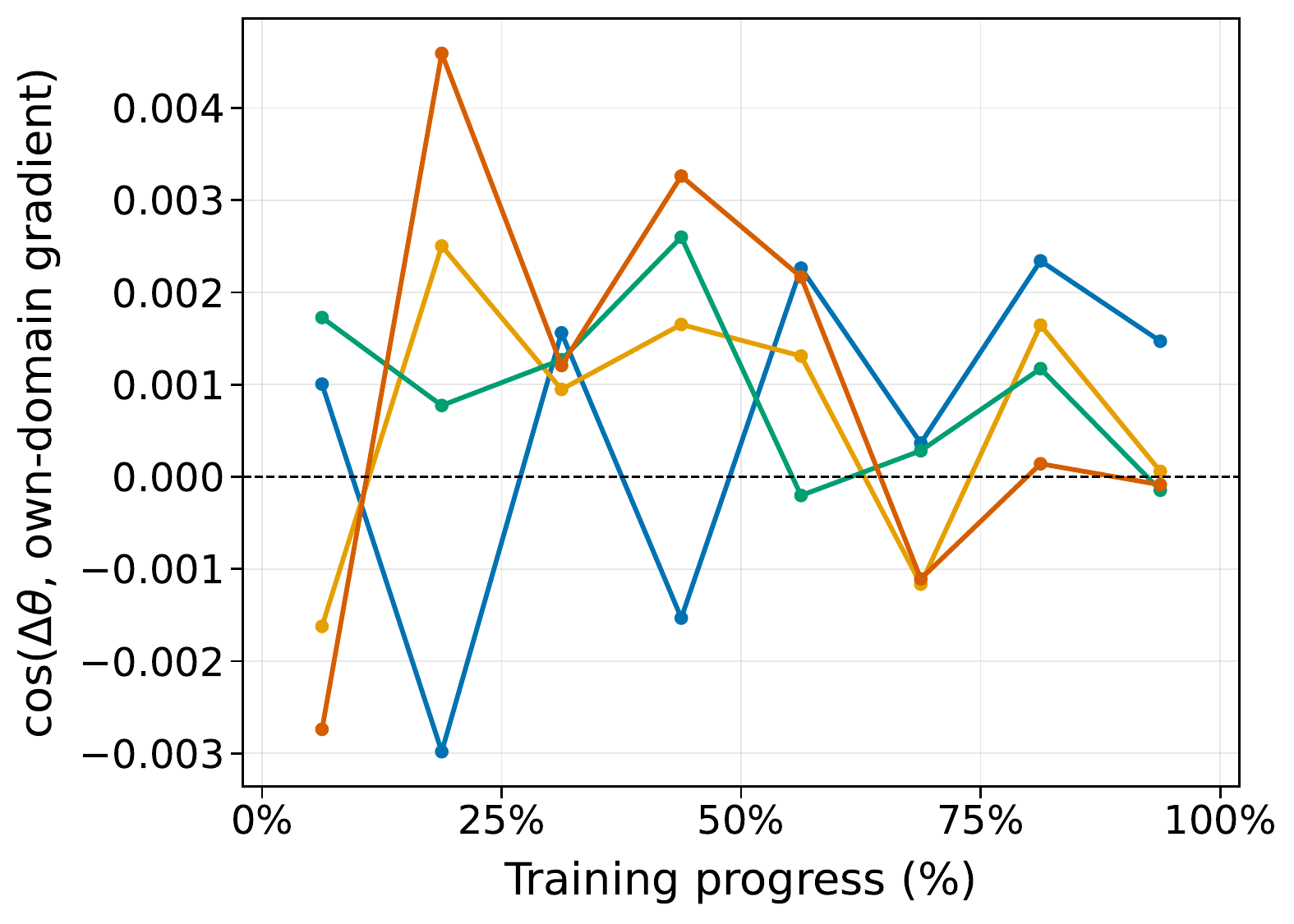}
        \caption{Independent RL Alignment}
        \label{fig:independent_reward_alignment}
    \end{subfigure}

    \vspace{2mm}
    \includegraphics[width=\linewidth]{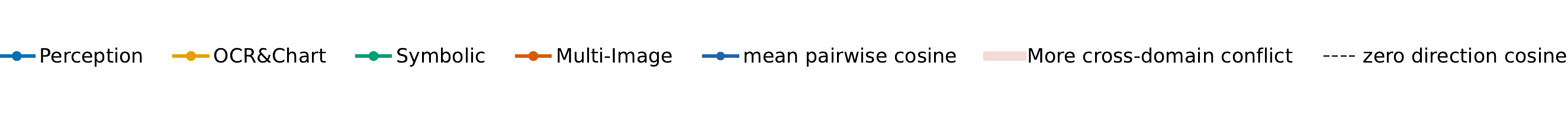}
    \vspace{-2mm}

    \caption{Reward-based update alignment in Mix-RL and independent capability RL. (a) Alignment of shared Mix-RL updates with capability-specific optimization directions. (b) Alignment of independently trained experts with their respective capability directions.}
    \label{fig:mixrl_independent_alignment}
\end{figure}

As shown in Figure~\ref{fig:mixrl_reward_alignment}, the shared Mix-RL update exhibits divergent alignment across capabilities, advancing along the optimization directions of some capabilities while opposing those of others. The recurring negative pairwise similarities further confirm cross-capability gradient interference. In contrast, Figure~\ref{fig:independent_reward_alignment} shows that independently trained capability experts maintain a positively aligned trajectory on average, despite occasional local fluctuations. Together, these observations suggest that shared-policy optimization introduces competing capability-specific update directions, potentially hindering specialized capability acquisition.
\begin{figure}[t]
    \centering

    \begin{subfigure}[t]{0.345\linewidth}
        \centering
        \includegraphics[width=\linewidth,trim=8 4 8 2,clip]{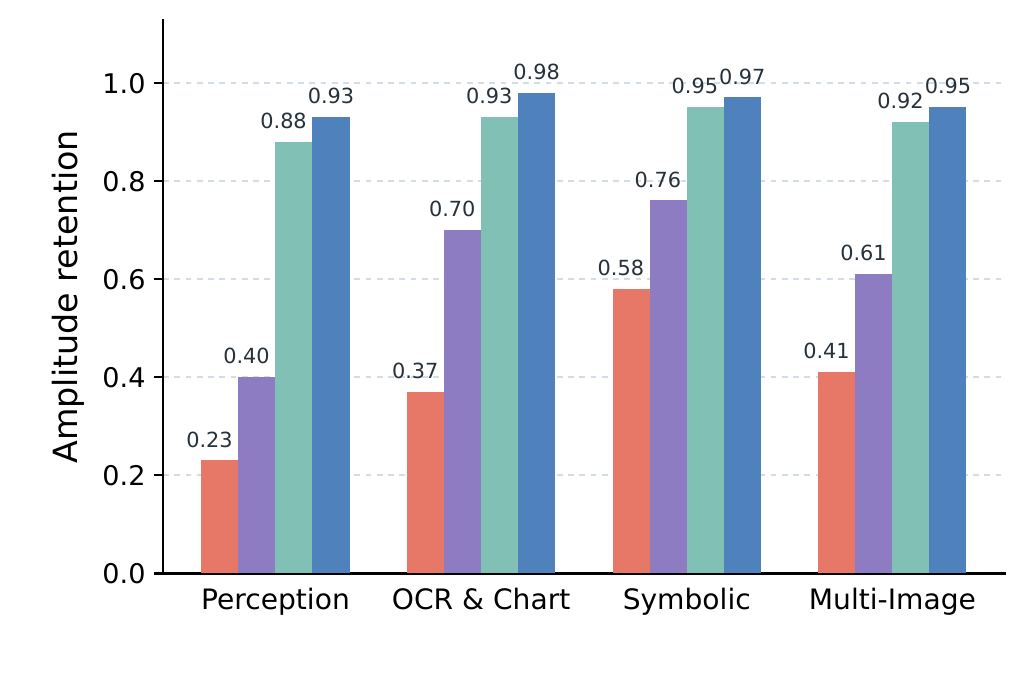}
        \caption{Expert Capability Retention}
        \label{fig:skill_expert_retention_amp}
    \end{subfigure}
    \hspace{-2mm}
    \begin{subfigure}[t]{0.345\linewidth}
        \centering
        \includegraphics[width=\linewidth,trim=8 4 8 2,clip]{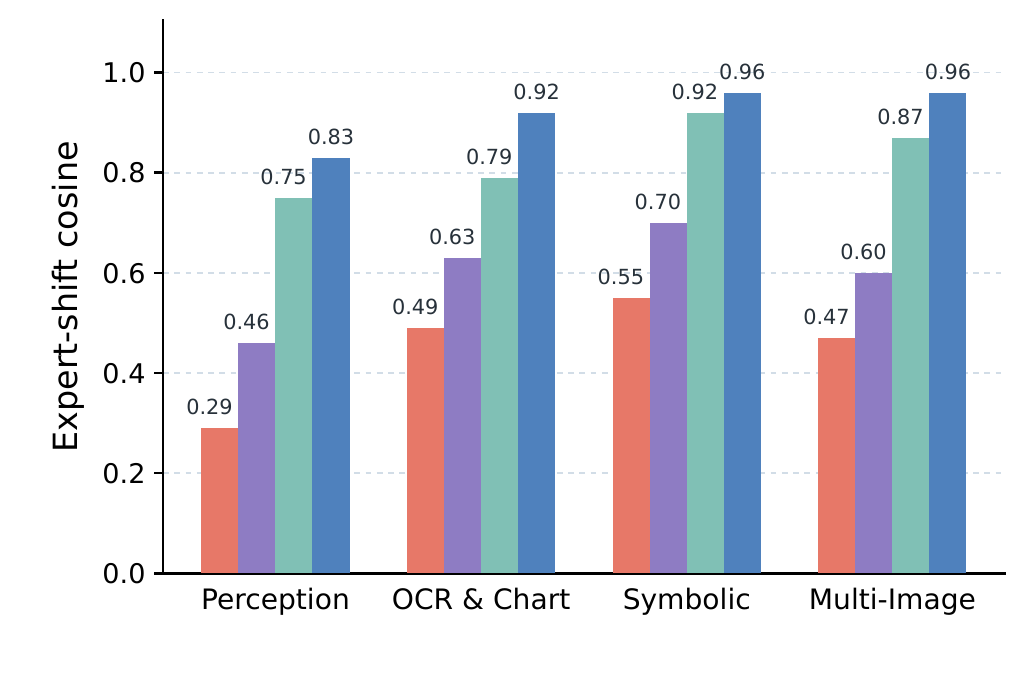}
        \caption{Expert Capability Alignment}
        \label{fig:skill_expert_retention_cos}
    \end{subfigure}
    \hspace{-2mm}
    \begin{subfigure}[t]{0.285\linewidth}
        \centering
        \includegraphics[width=\linewidth,trim=6 4 6 2,clip]{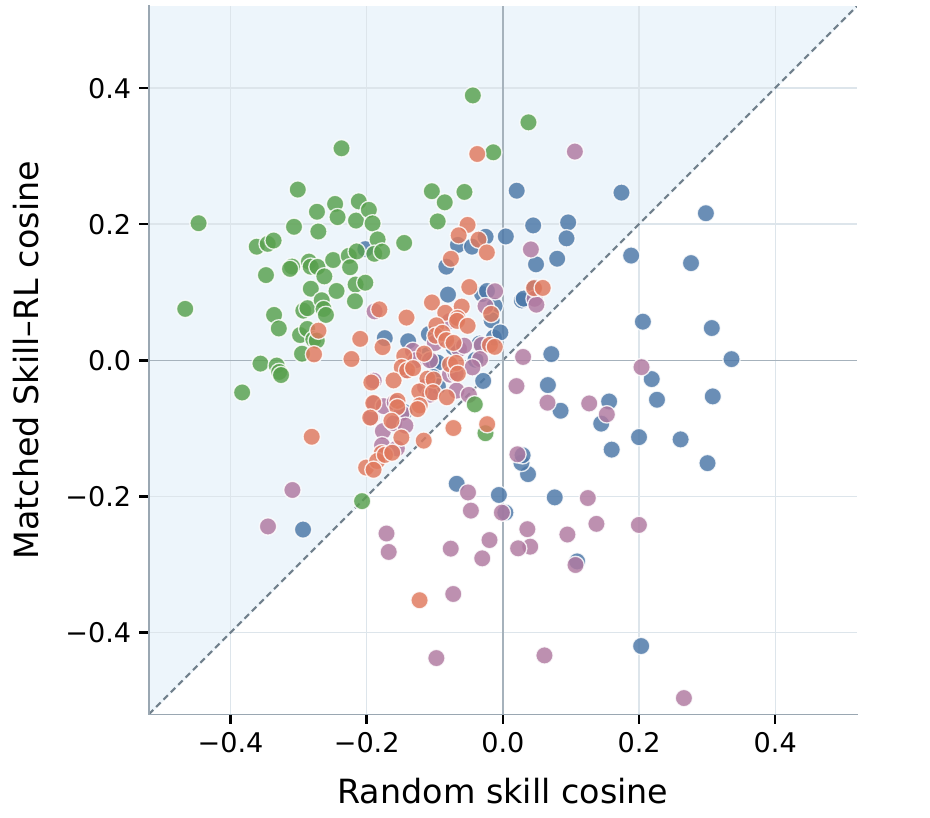}
        \caption{Matched vs. Random Skill}
        \label{fig:skill_expert_retention_scatter}
    \end{subfigure}

    \vspace{1.2mm}

    \includegraphics[width=0.86\linewidth]{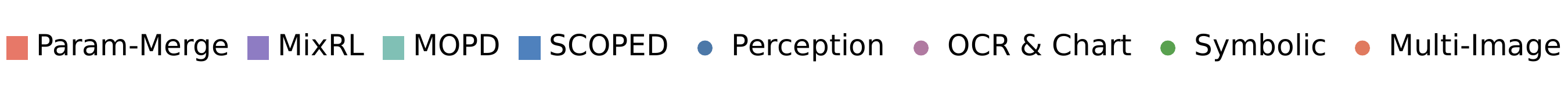}

    \vspace{-1mm}

    \caption{Effectiveness of skill-decoupled expert supervision. SCOPED achieves stronger domain expert capability retention and alignment than alternative integration methods, while matched domain skills provide more expert-aligned guidance than random skills.}
    \label{fig:skill_expert_retention}
\end{figure}
\subsection{Details of Expert-Retention Analysis}
\label{app:expert_retention}

We provide the computation protocol for the expert-retention analysis in Figure~\ref{fig:skill_retention_alignment}. For each capability, the independently trained capability expert without additional skill conditioning is used as the reference. All comparisons are performed using fixed-trajectory teacher forcing so that different unified models are evaluated under identical response states.

For each input $x_i$, the base policy $\pi_{\mathrm{base}}$ greedily generates a trajectory
\begin{equation}
\tau_i=(\tau_{i,1},\ldots,\tau_{i,T_i}),\qquad
h_{i,t}=(x_i,\tau_{i,<t}),
\label{eq:fixed_trajectory_states}
\end{equation}
and all compared policies are evaluated on the same states $\{h_{i,t}\}_{t=1}^{T_i}$. This avoids confounding functional differences with changes in generated trajectories.

\paragraph{Functional Expert Retention.}
For any policy $\pi$, we represent its functional deviation from the base policy along $\tau_i$ using the observed-token log-probability changes:
\begin{equation}
\mathbf{a}_{\pi}^{(i)}=\left(\log \pi(\tau_{i,t}\mid h_{i,t})-\log \pi_{\mathrm{base}}(\tau_{i,t}\mid h_{i,t})
\right)_{t=1}^{T_i}.
\label{eq:expert_retention_token_shift}
\end{equation}
Let $\mathbf{a}_{c}^{(i)}$ denote the vector produced by the corresponding capability expert $\pi_{\theta_c}$, and let $\mathbf{a}_{\theta}^{(i)}$ denote that of a unified model $\pi_{\theta}$.

We measure directional agreement with the capability expert using cosine similarity:
\begin{equation}
C_i(\theta,c)=\frac{\left\langle\mathbf{a}_{\theta}^{(i)},\mathbf{a}_{c}^{(i)}\right\rangle}{\left\|\mathbf{a}_{\theta}^{(i)}\right\|_2\left\|\mathbf{a}_{c}^{(i)}\right\|_2}.
\label{eq:expert_retention_direction}
\end{equation}
A larger $C_i$ indicates that the functional change induced by the unified model is more closely aligned with that of the corresponding capability expert.

We further measure the projected amplitude retained along the expert direction:
\begin{equation}
R_i(\theta,c)=\frac{\left\langle\mathbf{a}_{\theta}^{(i)},\mathbf{a}_{c}^{(i)}\right\rangle}{\left\|\mathbf{a}_{c}^{(i)}\right\|_2^2
}.
\label{eq:expert_retention_amplitude}
\end{equation}
Here, $R_i=1$ indicates that the unified model preserves the same projected magnitude as the expert, $0<R_i<1$ indicates attenuation, $R_i>1$ indicates amplification, and $R_i<0$ indicates an opposing component.

Both metrics are first averaged over samples within each capability and then macro-averaged across the four capabilities to obtain the results reported in Figure~\ref{fig:skill_retention_alignment}. These metrics quantify functional agreement with independently trained capability experts under identical trajectory states rather than directly measuring downstream task accuracy.

\subsection{Capability-Wise Expert Retention and Skill Alignment}

To complement the macro-averaged results presented in the main text, we further examine expert capability retention and alignment across individual capabilities. As shown in Figure~\ref{fig:skill_expert_retention_amp} and Figure~\ref{fig:skill_expert_retention_cos}, \method{} consistently achieves higher amplitude retention and expert alignment across all four capabilities, demonstrating more effective preservation of specialized expert behavior.

To examine whether these improvements arise from capability-specific guidance rather than additional contextual information, we compare matched capability skills with randomly assigned skills. As shown in Figure~\ref{fig:skill_expert_retention_scatter}, most samples lie above the diagonal, indicating that matched skills induce functional changes more closely aligned with the corresponding expert directions than random skills. These results suggest that the extracted skills provide targeted guidance consistent with capability-specific expertise.

\subsection{Qualitative Analysis of Skill-Guided Visual Attention}
\label{sec:skill_attention_case}

To complement the quantitative attention analysis in Sec.~\ref{sec:skill_expert_retention}, we further visualize how skill conditioning affects the visual attention of capability experts on a representative example.

\begin{figure}[t]
    \centering
    \setlength{\tabcolsep}{1pt}

    \resizebox{0.98\linewidth}{!}{%
    \begin{tabular}{@{}cccc@{}}
        \includegraphics[
            height=3.0cm,
            trim=0 0 50 0,
            clip
        ]{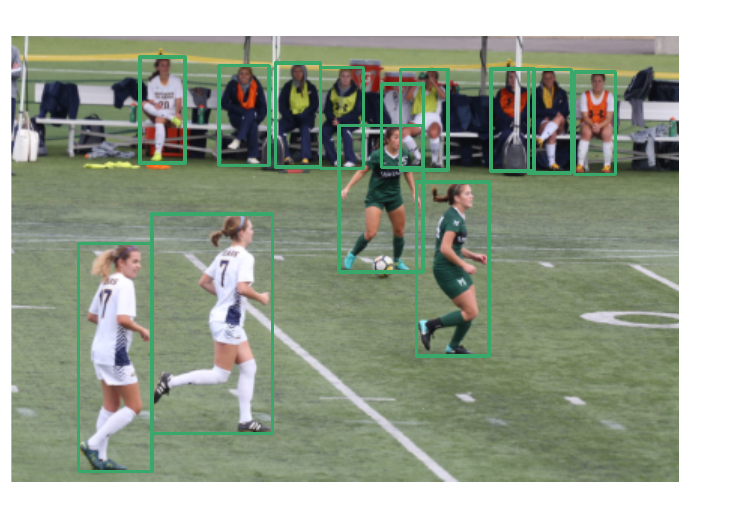}
        &
        \includegraphics[
            height=3.0cm,
            trim=0 0 45 0,
            clip
        ]{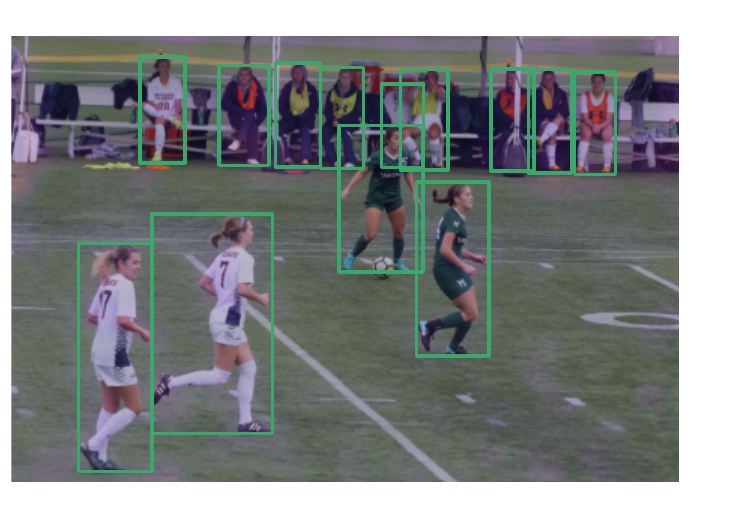}
        &
        \includegraphics[
            height=3.0cm
        ]{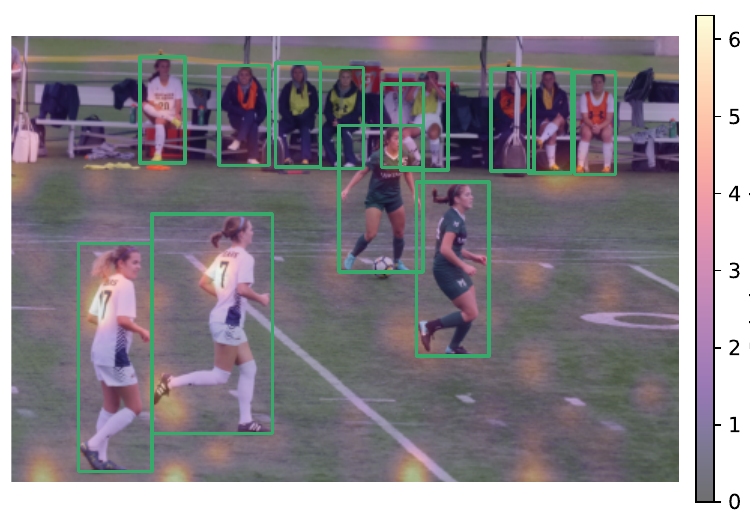}
        &
        \includegraphics[
            height=3.0cm
        ]{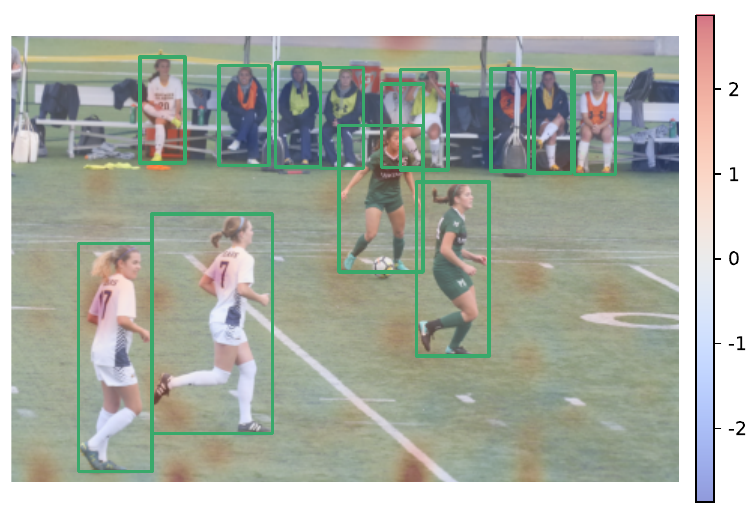}
        \\[-1mm]

        {\small (a) Target objects}
        &
        {\small (b) Baseline attention}
        &
        {\small (c) Skill-conditioned attention}
        &
        {\small (d) Skill-induced attention change}
    \end{tabular}%
    }

    \caption{Qualitative visualization of skill-induced attention reallocation. Green boxes indicate target objects. Red and blue in the attention-change map indicate increased and decreased attention after skill conditioning.}
    \label{fig:q799_attention_case}
\end{figure}

As illustrated in Figure~\ref{fig:q799_attention_case}, we compare the same capability expert with and without its matched skill under an identical input. Without skill conditioning, the expert distributes substantial attention over both task-relevant and irrelevant visual regions. In contrast, conditioning on the matched skill shifts more attention toward the visual evidence relevant to the current task, as further highlighted by the attention-difference map. This qualitative example is consistent with the layer-wise attention analysis in Figure~\ref{fig:attn_hit_layerwise}, providing an intuitive illustration of how skill conditioning can regularize expert visual attention toward task-relevant evidence.

\subsection{Case Studies}
\label{app:case_studies}

\begin{figure}[t]
    \centering
    \includegraphics[width=\linewidth]{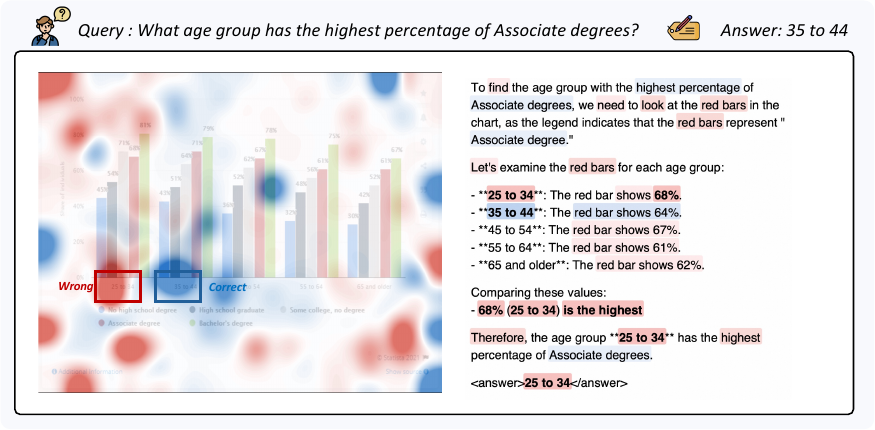}
\caption{
Qualitative effect of skill-augmented expert supervision.
Compared with \method{} w/o Skill Augmentation, \method{} strengthens responses supporting the correct answer while suppressing those associated with the erroneous trajectory. Blue indicates enhanced responses, while red indicates suppressed responses.
}
    \label{fig:case_study}
\end{figure}

To better understand how skill-augmented distillation changes the behavior of the unified model, we present a representative example in Figure~\ref{fig:case_study}. The task requires identifying the age group with the highest percentage of Associate degrees. While \method{} w/o Skill Augmentation follows an erroneous reasoning trajectory and predicts ``25 to 34'', the full \method{} model correctly identifies ``35 to 44''.

The visualization further illustrates how skill-augmented supervision affects the model's use of visual evidence and token-level predictions. Compared with \method{} w/o Skill Augmentation, \method{} allocates more attention to the region supporting the correct answer while reducing attention to the misleading region associated with the incorrect prediction. At the token level, \method{} also suppresses responses along the erroneous trajectory and strengthens those supporting the correct prediction.

This case complements the quantitative analyses in Section~\ref{sec:skill_expert_retention}, showing that skill-augmented expert supervision can lead the unified model to rely more on task-relevant visual evidence and avoid erroneous reasoning patterns during generation.

We further present representative examples from PixMo-Count, MUIRBench, and MMSI-Bench to examine whether these improvements are reflected across different visual capabilities. On PixMo-Count, the Qwen3-VL-Instruct backbone overlooks one aircraft and predicts six, whereas \method{} systematically scans and verifies individual instances before correctly identifying all seven aircraft. On MUIRBench, the backbone is distracted by appearance differences between maps, while \method{} instead compares stable structural and spatial cues and correctly identifies the corresponding image. Similarly, on MMSI-Bench, \method{} tracks the relative displacement of persistent visual landmarks across images and correctly infers the movement direction, whereas the backbone reaches an incorrect conclusion from a coarser comparison of the scenes.

\begin{figure}[t]
\centering
\includegraphics[width=\linewidth]{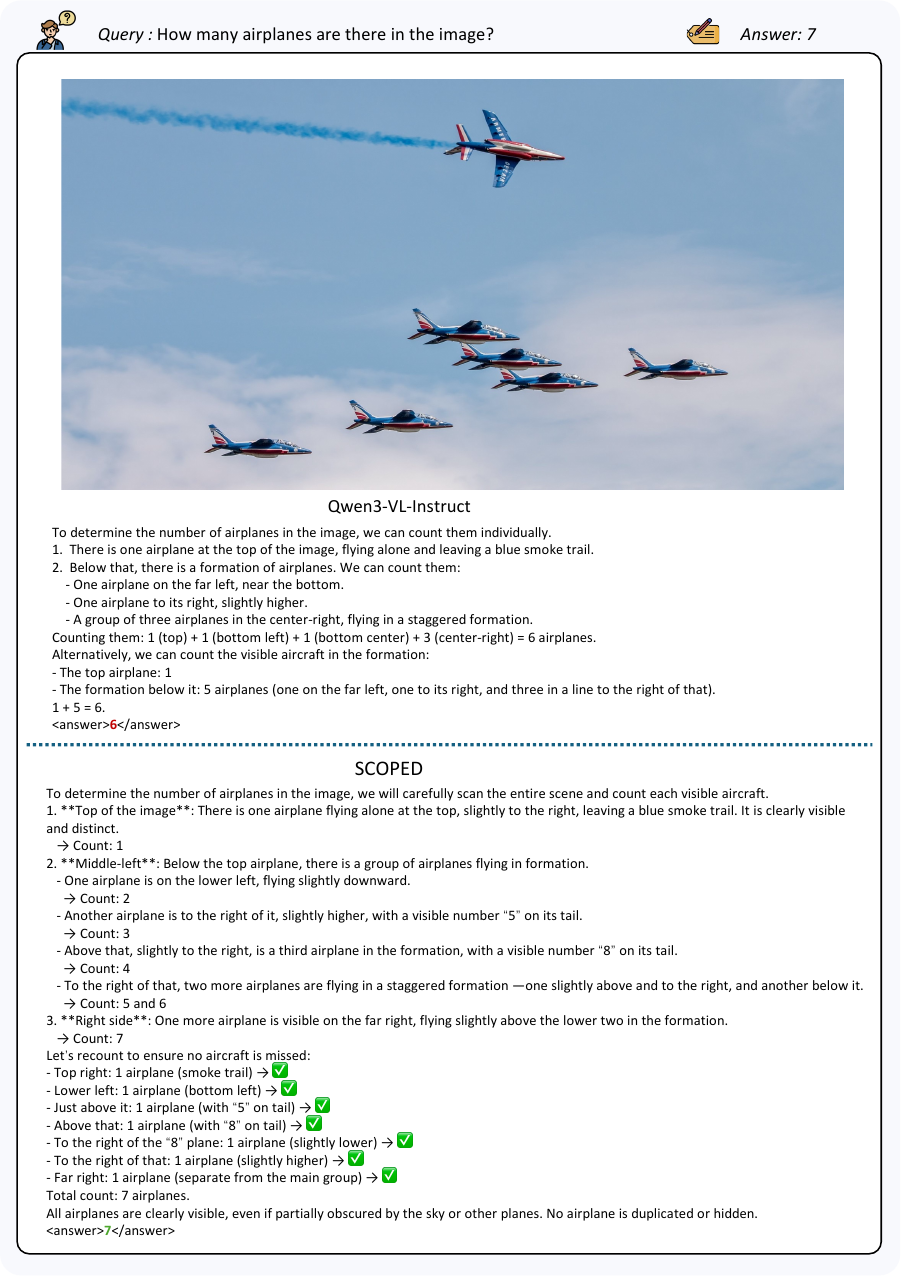}
\caption{Case study of PixMo-Count.}
\label{fig:case1}
\end{figure}

\begin{figure}[t]
\centering
\includegraphics[width=\linewidth]{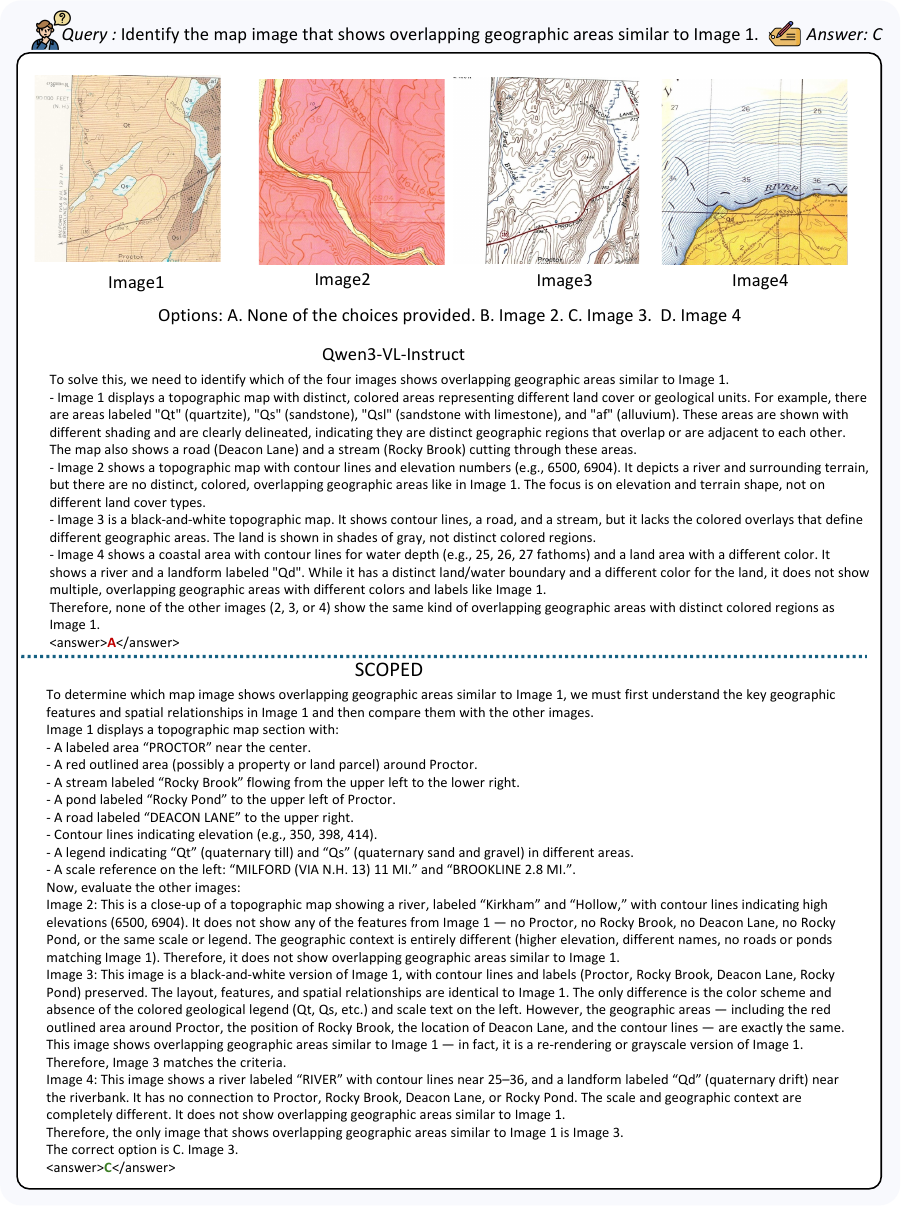}
\caption{Case study of MUIRBench.}
\label{fig:case2}
\end{figure}

\begin{figure}[t]
\centering
\includegraphics[width=\linewidth]{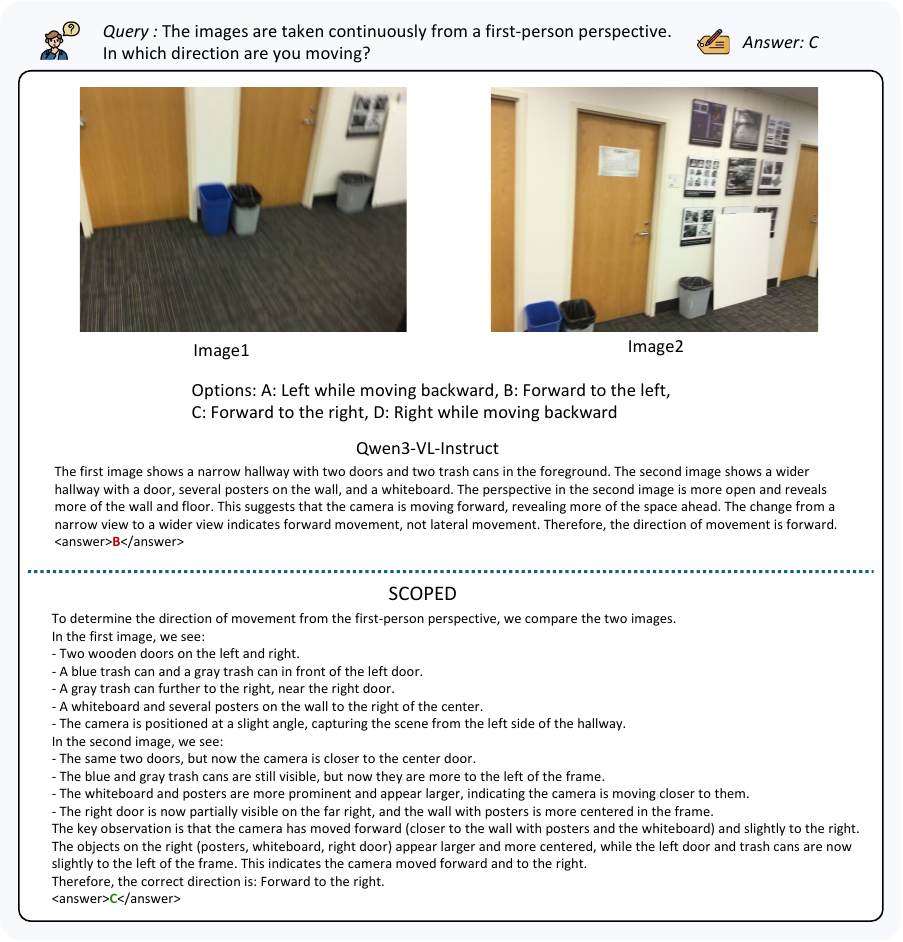}
\caption{Case study of MMSI-Bench.}
\label{fig:case3}
\end{figure}

Together, these examples illustrate that the gains of \method{} extend across fine-grained perception, cross-image correspondence, and multi-image spatial reasoning. Rather than relying on superficial visual patterns, the unified model exhibits more systematic evidence aggregation and verification, consistent with the quantitative improvements observed across the evaluated capability groups.
\FloatBarrier

\subsection{Prompt Examples}
\label{app:prompt_examples}

We provide the prompt templates used for both skill construction and evaluation.

\paragraph{Skill Construction Prompts.}
Our skill construction procedure uses three main prompts. First, the successful-trajectory self-reflection prompt asks the optimized capability expert to identify reusable verification behaviors and their activation conditions from its successful trajectories. Second, the failed-trajectory self-reflection prompt asks the same expert to diagnose causal failure mechanisms and derive corresponding corrective principles. Finally, the evidence-grounded skill induction prompt provides the curated evidence ledger to the same expert and asks it to consolidate recurring success and failure patterns into a compact capability-level skill. The corresponding prompt templates are shown in Figures~\ref{fig:success_reflection_prompt}--\ref{fig:skill_induction_prompt}.

\paragraph{Evaluation Prompts.}
Figure~\ref{fig:vanilla_prompt} presents the vanilla evaluation prompt, where the model receives the input image(s), the benchmark-specific query, and the corresponding response instruction. The model is instructed to reason step by step and place its final response within \texttt{\textless answer\textgreater} and \texttt{\textless/answer\textgreater} tags. The required output format is instantiated according to the benchmark: multiple-choice tasks require the final option letter, OCR tasks follow the official per-sample recognition instructions.

For evaluating skill-augmented capability experts in the ablation and analysis experiments, we preserve the original multimodal input and response instruction and condition each expert on its corresponding capability skill. The four extracted skills correspond to perception, OCR and chart understanding, symbolic reasoning, and multi-image reasoning, respectively. The unified student does not receive explicit skill conditioning during inference.

\clearpage

\begin{figure}[p]
\centering
\small
\begin{tcolorbox}[
promptbox,
title={Successful-Trajectory Self-Reflection Prompt}
]

\begin{tabularx}{\linewidth}{
@{}
>{\bfseries\raggedright\arraybackslash}p{0.22\linewidth}
Y
@{}
}
Role
&
You are an expert success-pattern analyst for \texttt{\{domain\}} tasks.
\\[2pt]

Input
&
A successful trajectory generated by a strong domain expert, together with the task query and ground-truth target.
\\[2pt]

Objective
&
Identify a lightweight verification or self-check principle that contributed to the successful response.
\\[2pt]

Analysis
&
Identify (1) the effective self-check or verification behavior,
(2) the subtle cue that could otherwise cause an error, and
(3) the condition under which this check should be applied.
\\[2pt]

Constraints
&
Do not prescribe a step-by-step solving procedure, force a chain-of-thought format, copy instance-specific answers, or introduce dataset-specific details. Prefer conditional verification guidance that is activated only when necessary.
\end{tabularx}

\tcblower

\textbf{Required Output}
\vspace{1mm}

\begin{tabularx}{\linewidth}{
@{}
>{\bfseries\raggedright\arraybackslash}p{0.22\linewidth}
Y
@{}
}
Format
&
Return a valid JSON object containing a concise reusable skill rule and its activation condition.
\\[2pt]

Task
&
\texttt{\{question\}}
\\[2pt]

Expert Trajectory
&
\texttt{\{trajectory\}}
\\[2pt]

Ground Truth
&
\texttt{\{ground\_truth\}}
\end{tabularx}

\end{tcolorbox}

\caption{Prompt used by the optimized capability expert to extract a reusable self-check principle from a successful trajectory.}
\label{fig:success_reflection_prompt}
\vspace{-4mm}
\begin{tcolorbox}[
promptbox,
title={Failed-Trajectory Self-Reflection Prompt}
]

\begin{tabularx}{\linewidth}{
@{}
>{\bfseries\raggedright\arraybackslash}p{0.22\linewidth}
Y
@{}
}
Role
&
You are an expert failure-analysis agent for \texttt{\{domain\}} tasks.
\\[2pt]

Input
&
A failed expert trajectory together with the task query, predicted answer, and ground-truth target.
\\[2pt]

Objective
&
Diagnose the underlying cause of failure and convert it into reusable corrective guidance.
\\[2pt]

Analysis
&
Identify (1) the mismatch between the prediction and ground truth,
(2) the point at which the reasoning failed,
(3) the underlying failure mechanism, and
(4) a general corrective principle and its activation condition.
\\[2pt]

Constraints
&
Derive general principles rather than instance-specific corrections. Do not copy sample-specific facts, dataset names, numerical values, or task-specific surface details.
\end{tabularx}

\tcblower

\textbf{Required Output}
\vspace{1mm}

\begin{tabularx}{\linewidth}{
@{}
>{\bfseries\raggedright\arraybackslash}p{0.22\linewidth}
Y
@{}
}
Format
&
Return a valid JSON object describing the failure cause, corrective rule, and condition under which the rule should be applied.
\\[2pt]

Task
&
\texttt{\{question\}}
\\[2pt]

Expert Trajectory
&
\texttt{\{trajectory\}}
\\[2pt]

Final Answer
&
\texttt{\{prediction\}}
\\[2pt]

Ground Truth
&
\texttt{\{ground\_truth\}}
\end{tabularx}

\end{tcolorbox}

\caption{Prompt used by the optimized capability expert to diagnose a failed trajectory and extract reusable corrective guidance.}
\label{fig:failure_reflection_prompt}
\end{figure}

\clearpage

\clearpage

\begin{figure}[p]
\centering
\small

\begin{tcolorbox}[
promptbox,
title={Evidence-Grounded Capability Skill Induction Prompt}
]

\begin{tabularx}{\linewidth}{
@{}
>{\bfseries\raggedright\arraybackslash}p{0.22\linewidth}
Y
@{}
}
Role
&
You are improving an expert vision-language model using evidence distilled from its own training trajectories.
\\[2pt]

Input
&
\texttt{\{Evidence Ledgers\}} containing successful and failed trajectory patterns for the current domain.
\\[2pt]

Domain Guidance
&
\texttt{\{Domain-Specific Guidance\}} specifying how the observed failure mechanisms should transfer across tasks within the domain.
\\[2pt]

Objective
&
Convert recurring success and failure patterns into a compact domain-level skill that provides reusable verification and corrective guidance.
\\[2pt]

Rule Form
&
Each task-specific rule should follow:
\textit{``When [specific trigger], [verification or corrective action].''}
\\[2pt]

Task Routing
&
Organize rules by task section so that only the checks relevant to the current question type are applied.
\\[2pt]

Grounding
&
Ground each task section in supporting failure evidence and, when available, contrasting successful evidence.
\\[2pt]

Constraints
&
Do not copy instance-specific answers, values, coordinates, or memorized examples.
Avoid unsupported rules, unnecessary step-by-step procedures, and instructions that suppress reasoning.
Prefer a small number of precise, conditionally activated rules.
\end{tabularx}

\tcblower

\textbf{Required Output}
\vspace{1mm}

\begin{tabularx}{\linewidth}{
@{}
>{\bfseries\raggedright\arraybackslash}p{0.22\linewidth}
Y
@{}
}
Objective
&
A concise objective for the domain skill.
\\[2pt]

Evidence Protocol
&
A small set of shared rules applicable across tasks in the domain.
\\[2pt]

Task-Specific Checks
&
For each task section, a small set of trigger-conditioned verification or corrective rules.
\\[2pt]

Final Verification
&
A small set of final checks applied before producing the answer.
\end{tabularx}

\end{tcolorbox}

\caption{Prompt used by the optimized capability expert to induce a domain-level skill from curated trajectory-derived evidence. The resulting skill contains shared guidance together with conditionally activated task-specific checks.}
\label{fig:skill_induction_prompt}

\vspace{-4mm}

\begin{tcolorbox}[
promptbox,
title={Vanilla Evaluation Prompt}
]

\begin{tabularx}{\linewidth}{
@{}
>{\bfseries\raggedright\arraybackslash}p{0.22\linewidth}
Y
@{}
}
Input
&
\{\texttt{Input Image(s)}\}
\\[2pt]

Task Query
&
\{\texttt{Dataset-Specific Question}\}
\\[2pt]

Response Instruction
&
Please reason step by step, and put your final answer within
\answeropen{} and \answerclose{}.
\end{tabularx}

\tcblower

\textbf{Task-Specific Output Formats}
\vspace{1mm}

\begin{tabularx}{\linewidth}{
@{}
>{\bfseries\raggedright\arraybackslash}p{0.22\linewidth}
Y
@{}
}
Multiple Choice
&
Put only the final option letter within
\answeropen{} and \answerclose{}.
\\[2pt]
OCR Recognition
&
Follow the official per-sample recognition instruction
provided by the benchmark.
\end{tabularx}

\end{tcolorbox}

\caption{Vanilla prompt template used for evaluation. The query and response format are instantiated for each benchmark.}
\label{fig:vanilla_prompt}

\end{figure}

\clearpage

\begin{figure*}[t]
\centering

\begin{tcolorbox}[
promptbox,
title={Capability Skill: Fine-Grained Perception}
]

\small

\textbf{Objective}

Answer the current benchmark item accurately from the provided image(s) and question.

\vspace{2mm}
\textbf{Evidence Protocol}

\begin{itemize}
    \item Apply ONLY the task section matching the current question type; ignore the others.
    \item When the task involves sequential operations, update the working state after each step before proceeding.
    \item Do not use dataset names, filenames, memorized examples, or external information.
\end{itemize}

\textbf{Task-Specific Checks}

\textit{Apply when the question asks how many instances of a specific object/category appear.}

\begin{itemize}
    \item When a candidate instance is faint, translucent, partially cropped at the image edge, or only a tiny fragment is visible, exclude it from the count---only count objects that are clearly and substantially present.
    \item When the noun could refer to either a complete assembled item or its sub-components, count the most natural standalone unit that the noun refers to in everyday language.
\end{itemize}

\textit{Final verification:}
\begin{itemize}
    \item When the task involved subtraction or modification, confirm the final result reflects all operations applied in order.
    \item When your count includes a faint, translucent, or barely visible instance and you are uncertain, re-examine whether that instance is real before including it.
\end{itemize}

\textit{Apply when counting objects in natural photographs.}

\begin{itemize}
    \item When counting in a natural photograph, recount the entire scene after your first read---the dominant error is omission of partially hidden or edge-proximal instances.
    \item When spatial density or layout ambiguity arises, check for hidden or partial objects---include partially visible objects if context and boundaries support that they are real instances.
    \item When the count feels complete, do one final sweep of low-contrast and peripheral areas.
    \item When the target category could be counted as a whole set or as individual pieces, decide which unit the question asks for before counting and apply it consistently.
    \item When your first pass finds zero instances, run a second pass assuming the category may appear in an unusual, stylized, or partially occluded form before concluding it is absent.
\end{itemize}

\textit{Apply when locating a single object described by a referring expression.}

\begin{itemize}
    \item Locate exactly one object matching the full phrase. Interpret spatial terms from the viewer's perspective.
    \item When multiple same-category candidates exist, verify that the chosen object satisfies every word of the referring expression, especially spatial qualifiers.
    \item For ordinal expressions such as ``2nd from left,'' count from the edge explicitly named in the phrase.
\end{itemize}

\textbf{Output}

Follow the output format stated in the question exactly, including any answer tags it specifies.
Do not shorten, skip, or reorder the reasoning the question asks for, and never state the final answer before the checks above are done.

\end{tcolorbox}

\caption{Extracted Capability skill for fine-grained perception.}
\label{fig:skill_perception}
\end{figure*}
\begin{figure*}[t]
\centering

\begin{tcolorbox}[
promptbox,
title={Capability Skill: OCR and Chart Understanding}
]

\small

\textbf{Objective}

Answer the current benchmark item accurately from the provided image(s) and question.

\vspace{2mm}
\textbf{Evidence Protocol}

\begin{itemize}
    \item Apply ONLY the task section matching the current question type; ignore the others.
    \item When the task involves sequential operations, update the working state after each step before proceeding.
    \item Do not use dataset names, filenames, memorized examples, or external information.
\end{itemize}

\textbf{Task-Specific Checks}

\textit{Apply when answering questions about charts.}

\begin{itemize}
    \item When reading values from a chart, align with the nearest labeled gridline or tick mark---do not estimate between lines.
    \item When comparing two values, confirm whether the question asks for an absolute difference or a ratio.
    \item When a chart has dual axes or multiple series, verify that the value is read from the correct axis and series.
    \item When identifying a peak or extremum, cross-reference its position with both axes to confirm the exact value.
\end{itemize}

\textit{Final verification:}
\begin{itemize}
    \item When the answer's magnitude seems implausible given the chart's visual scale, re-read the value against the axis markings.
    \item When transcribing text, verify that no visible text region was skipped by scanning page edges and corners.
\end{itemize}

\textit{Apply when answering questions about presentation slide decks.}

\begin{itemize}
    \item When extracting data from a multi-page document, verify that the answer comes from the page matching the question's context rather than a visually similar page.
    \item When the question asks about a relationship or flow, trace connecting arrows or lines in the diagram before answering.
    \item When listing categories from a chart or table, ensure that all items are captured.
\end{itemize}

\textit{Final verification:}
\begin{itemize}
    \item When the answer's magnitude seems implausible given the visual scale, re-check the relevant values.
    \item When transcribing text, verify that no visible text region was skipped.
\end{itemize}

\textit{Apply when the task is pure text recognition/OCR. Activate only for long multi-line text, multi-column layouts, or peripheral text blocks that could be overlooked.}

\begin{itemize}
    \item Scan every region systematically, including the main body, headers, footers, captions, footnotes, watermarks, stamps, and table cells.
    \item For multi-column or multi-region layouts, transcribe each region in natural reading order and preserve completeness.
    \item For long CJK text, read each line fully before moving to the next line; do not merge lines or skip characters.
    \item When uncertain whether a faint or partial region contains real text, prefer preserving potentially valid text rather than omitting an entire region.
\end{itemize}

\textit{Final verification:}
\begin{itemize}
    \item Re-scan the four edges and every column or region of the image and confirm that no visible text line was skipped or truncated.
\end{itemize}

\textbf{Output}

Follow the output format stated in the question exactly, including any answer tags it specifies.
Do not shorten, skip, or reorder the reasoning the question asks for, and never state the final answer before the checks above are done.

\end{tcolorbox}

\caption{Extracted Capability skill for OCR, document, and chart understanding.}
\label{fig:skill_ocr_chart}
\end{figure*}
\begin{figure*}[t]
\centering

\begin{tcolorbox}[
promptbox,
title={Capability Skill: Multimodal Symbolic Reasoning}
]

\small

\textbf{Objective}

Answer the current benchmark item accurately from the provided image(s) and question.

\vspace{2mm}
\textbf{Evidence Protocol}

\begin{itemize}
    \item Apply ONLY the task section matching the current question type; ignore the others.
    \item When the task involves sequential operations, update the working state after each step before proceeding.
    \item Do not use dataset names, filenames, memorized examples, or external information.
\end{itemize}

\textbf{Task-Specific Checks}

\textit{Apply when solving geometry problems with diagrams.}

\begin{itemize}
    \item When a derivation yields a result, verify that it is consistent with the diagram's visible proportions; if the computed length or angle is implausible, re-check the derivation and arithmetic.
    \item For multi-step derivations, solve each intermediate quantity explicitly and carry it forward to avoid propagating arithmetic errors.
\end{itemize}

\textit{Final verification:}
\begin{itemize}
    \item Confirm that the answer is a bare numeric value, with no option letter or units appended.
\end{itemize}

\textit{Apply when solving visual pattern or inductive puzzles.}

\begin{itemize}
    \item For inductive pattern or sequence puzzles, identify the simplest visual rule consistent with all given figures---such as rotation, color or shape cycles, or element addition/removal---and apply it forward.
    \item For other puzzle types, rely on direct visual inspection or straightforward reasoning rather than unnecessarily applying pattern-detection rules.
\end{itemize}

\textit{Apply when solving mathematical word problems requiring multi-step computation.}

\begin{itemize}
    \item When choosing among options, compute the answer independently first and then match it to the corresponding option; do not let the answer choices bias the derivation.
\end{itemize}

\textit{Final verification:}
\begin{itemize}
    \item Confirm that the final answer is an option letter (A, B, C, or D).
\end{itemize}

\textbf{Output}

Follow the output format stated in the question exactly, including any answer tags it specifies.
Do not shorten, skip, or reorder the reasoning the question asks for, and never state the final answer before the checks above are done.

\end{tcolorbox}

\caption{Extracted Capability skill for multimodal symbolic reasoning.}
\label{fig:skill_symbolic}
\end{figure*}
\begin{figure*}[t]
\centering

\begin{tcolorbox}[
promptbox,
title={Capability Skill: Multi-Image Reasoning}
]

\small

\textbf{Objective}

Answer the current benchmark item accurately from the provided image(s) and question.

\vspace{2mm}
\textbf{Evidence Protocol}

\begin{itemize}
    \item Apply ONLY the task section matching the current question type; ignore the others.
    \item When the task involves sequential operations, update the working state after each step before proceeding.
    \item Do not use dataset names, filenames, memorized examples, or external information.
\end{itemize}

\textbf{Task-Specific Checks}

\textit{Apply when answering questions that reference multiple images.}

\begin{itemize}
    \item When comparing images for thematic or semantic similarity, verify that at least two distinct, non-redundant visual anchors are consistent across candidate images.
    \item When evaluating whether a collection of images satisfies a condition, confirm the relevant quantifier: a universal claim fails if any required image contradicts it, whereas an existential claim requires only one valid instance.
    \item When identifying a paired comparison, verify that the images share a meaningful semantic or structural relationship rather than superficial visual resemblance.
    \item When no candidate image matches the reference in key structural or spatial anchors, do not force a comparison.
    \item When judging whether images share the same scene, building, or identity, prioritize stable structural anchors over decorative or transient elements.
\end{itemize}

\textit{Final verification:}
\begin{itemize}
    \item Confirm that the selected answer is uniquely supported by the visual evidence.
    \item Do not infer absence solely because an object or motion is not clearly visible.
    \item For directional or quantifier reasoning, verify that the correct reference frame and logical interpretation have been used.
\end{itemize}

\textit{Apply for multi-image spatial and relational reasoning questions.}

\begin{itemize}
    \item When evaluating directional terms such as ``turned to the right'' or ``facing left,'' confirm the intended reference frame rather than relying on image tilt or an object's internal orientation.
    \item When a statement specifies a spatial relationship, interpret it literally unless the question explicitly introduces another reference frame.
    \item When a ``cannot be determined'' option is available and a precise answer requires information that is not visually recoverable, prefer the insufficient-information option rather than inferring unsupported details.
\end{itemize}

\textit{Final verification:}
\begin{itemize}
    \item Confirm that the selected answer is uniquely supported by the visual evidence.
    \item Avoid reasoning based on inferred absence when the relevant evidence is not visible.
    \item Verify the interpretation of directional and quantifier relations before producing the final answer.
\end{itemize}

\textbf{Output}

Follow the output format stated in the question exactly, including any answer tags it specifies.
Do not shorten, skip, or reorder the reasoning the question asks for, and never state the final answer before the checks above are done.

\end{tcolorbox}

\caption{Extracted Capability skill for multi-image reasoning.}
\label{fig:skill_multi_image}
\end{figure*}
\end{document}